%% file: main.tex
\documentclass[10pt, conference]{IEEEtran}
\IEEEoverridecommandlockouts
\usepackage{cite}
\usepackage{amsmath,amssymb,amsfonts}
\usepackage{algorithmic}
\usepackage{graphicx}
\usepackage{textcomp}
\usepackage{xcolor}
\usepackage{booktabs}
\usepackage{multirow}
\usepackage{multicol}

\def\BibTeX{{\rm B\kern-.05em{\sc i\kern-.025em b}\kern-.08em
    T\kern-.1667em\lower.7ex\hbox{E}\kern-.125emX}}

\newcommand{\risto}[1]{{\color{purple}{\small\bf\sf [Risto: #1]}}}

\newcommand{\Skip}[1]{{}}

\newcommand{\etal}{\textit{et al}.\ }
\newcommand{\ie}{\textit{i}.\textit{e}.\ }
\newcommand{\eg}{\textit{e}.\textit{g}.\ }
\newcommand{\myfig}[1]{Figure \ref{#1}}
\newcommand{\mytable}[1]{Table \ref{#1}}

\DeclareMathOperator{\E}{\mathbb{E}}
\DeclareMathOperator*{\argmax}{arg\,max}

\title{Deep Reinforcement Learning for Multi-Driver Vehicle Dispatching and Repositioning Problem}

\begin{document}

\Skip{

  1. John Holler (Student) , University of Michigan, United States
Email: <johnholl@umich.edu>
* 2. Risto Vuorio, University of Michigan, United States
Email: <riv@umich.edu>
  3. Tiancheng Jin, University of Michigan, United States
Email: <jintia@umich.edu>
  4. Satinder Singh, University of Michigan, United States
Email: <baveja@umich.edu>
  5. Zhiwei Qin, Didi Research America, United States
Email: <qinzhiwei@didiglobal.com>
  6. Prof. Jieping Ye, Didi Chuxing, China
Email: <yejieping@didiglobal.com>
  7. Xiaocheng Tang, Didi Research America, United States
Email: <xiaochengtang@didiglobal.com>
  8. Yan Jiao, Didi Research America, United States
Email: <yanjiao@didiglobal.com>
  9. Chenxi Wang, Didi Research, China
Email: <wangchenxi@didiglobal.com>
}

\author{
    \IEEEauthorblockN{
        John Holler\IEEEauthorrefmark{1}\IEEEauthorrefmark{2},
        Risto Vuorio\IEEEauthorrefmark{1}\IEEEauthorrefmark{2},
        Zhiwei Qin\IEEEauthorrefmark{3},
        Xiaocheng Tang\IEEEauthorrefmark{3},
        Yan Jiao\IEEEauthorrefmark{3},
        Tiancheng Jin\IEEEauthorrefmark{2},
    }
    \IEEEauthorblockN{
        Satinder Singh\IEEEauthorrefmark{2},
        Chenxi Wang\IEEEauthorrefmark{3}
        and Jieping Ye\IEEEauthorrefmark{3}
    }
    \IEEEauthorblockA{
        \IEEEauthorrefmark{1}Equal Contribution,
        \IEEEauthorrefmark{2}University of Michigan,
        \IEEEauthorrefmark{3}Didi Chuxing \\
        \texttt{\{johnholl, riv, jintia, baveja\}@umich.edu} \\
        \texttt{\{qinzhiwei, xiaochengtang, yanjiao, wangchenxi, yejieping\}@didiglobal.com}
    }
}

\maketitle

\begin{abstract}
    Order dispatching and driver repositioning (also known as fleet management) in the face of spatially and temporally varying supply and demand are central to a ride-sharing platform marketplace.
    Hand-crafting heuristic solutions that account for the dynamics in these resource allocation problems is difficult, and may be better handled by
    an end-to-end machine learning method.
    Previous works have explored machine learning methods to the problem from a high-level perspective, where the learning method is responsible for either repositioning the drivers or dispatching orders,
    and as a further simplification, the drivers are considered independent agents maximizing their own reward functions.
    In this paper we present a deep reinforcement learning approach for tackling the full fleet management and dispatching problems.
    In addition to treating the drivers as individual agents, we consider the problem from a system-centric perspective,
    where a central fleet management agent is responsible for decision-making for all drivers.
\end{abstract}


\input{text/introduction.tex}
\input{text/related_work.tex}

\input{text/method.tex}
\input{text/experiments.tex}
\input{text/conclusion.tex}

\bibliography{references}
\bibliographystyle{IEEEtran}

\clearpage

\appendices
\input{text/appendix.tex}

\end{document}

%% file: text/introduction.tex
\section{Introduction}
\label{sec:introduction}
The order dispatching and fleet management system at a ride-sharing company must make decisions both
for assigning available drivers to nearby passengers (hereby called orders) and for repositioning
drivers who have no nearby orders.
These decisions have short-term effects on the revenue generated by the drivers and driver availability.
In the long term they change the distribution of drivers across the city, which 
in turn has a critical impact on how well future orders can be served.
Provident algorithmic solutions, which account for the short term and long-term
consequences of their decisions can improve the quality of service of the
ride-sharing platforms and are thus an important area of research.

Recent works~\cite{bello2016neural,nazari2018deep} have successfully applied Deep Reinforcement Learning (RL)
techniques to dispatching problems, such as the Traveling Salesman Problem (TSP) and the more general Vehicle
Routing Problem (VRP)~\cite{dantzig1959truck}, however they have primarily focused on {\em static} 
(\ie those where all orders are known up front) and/or {\em single-driver} dispatching problems.
In contrast to these problems, the fleet management and order dispatching problem ride-sharing platforms
face has multiple drivers and dynamically changing supply and demand conditions.
We refer to this dynamic dispatching and fleet management problem as the
Multi-Driver Vehicle Dispatching and Repositioning Problem (MDVDRP). 

VRPs and other problems similar to the MDVDRP are studied in the field of combinatorial optimization.
Exactly solving instances of these problems at the scale of real-world environment is computationally intractable \cite{toth2002vehicle}.
To deal with the intractability, heuristic solutions, which produce approximate solutions in polynomial time,
are often used~\cite{cordeau2002guide,pisinger2007general}.
The planning problem presented by the MDVDRP is related to the VRPs, but the complexity comes from the
dynamic nature of the assignment scenario rather than the intractability of computing the exact solution.
Drivers and orders appear in the assignment system at random points in time. In this dynamic assignment
setting, assignment decisions are made based on the current driver-order situation, without exact
information about future orders. A high performing assignment solution needs to account for unknown future
supply and demand conditions.

In realistic instances of the MDVDRP, the decision-making continues 24 hours a day and may involve thousands
of drivers and tens of thousands of customers.
Accounting for the spatially and temporally varying supply and demand conditions makes hand-crafting
heuristic solutions to these scenarios challenging.
In this paper, we explore a deep reinforcement learning approach to the MDVDRP.
The contributions of our work can be summarized as follows. (i) We introduce a new reinforcement
learning problem: the MDVDRP, which is motivated by the needs of real-world ride-sharing platforms.
(ii) We propose a novel network architecture for decision-making in a realistic problem
setting with variable sized observation and action spaces. (iii) We provide empirical analysis
of value-based and actor-critic methods on instances of the MDVDRP, including
instances based on real-world data.

%% file: text/related_work.tex
\section{Related Work}
\label{sec:related_work}
Recent machine learning approaches to dispatching and routing problems operate according to an encoding-decoding scheme,
where information is first processed into a fixed-sized representation, and then actions are decoded from this representation
\cite{bello2016neural,deudon2018learning}. Machine learning approaches to the more general problem family of combinatorial
optimization problems are surveyed in~\cite{bengio2018machine}.
Pointer networks~\cite{vinyals2015pointer} offer an approximate solution to traveling salesman problems by encoding cities
(in our terminology, orders) with a recurrent network, and then producing a solution by sequentially ``pointing'' to orders
using an attention mechanism~\cite{mnih2014recurrent}. The network is trained with a supervised loss function by imposing a
fixed ordering rule on decoded orders.
Bello \etal propose training an architecture similar to the pointer networks with  policy gradients instead of a
supervised loss. Using policy gradients allows them to dispense with the fixed ordering of the outputs during the decoding phase~\cite{bello2016neural}.
Similarly, we use reinforcement learning to train our networks. An alternative to pointer networks is explored
in~\cite{kool2019attention}, where graph convolutional networks are used for solving routing problems.
We follow an architecture related to~\cite{nazari2018deep},
which uses an attention mechanism for encoding the inputs.
We depart from their architecture in two ways. First, we replace the input attention layers with layers that compute their
output elementwise. Second, we remove the recurrent network used in the decoder.

In practice, order dispatching and fleet management problems are often solved with heuristic solutions,
which construct an approximation to the true problem by ignoring the spatial extent, or the temporal dynamics,
or both, and solve the approximate problem exactly.
One example of a heuristic solution to the order dispatching problem is the myopic pickup distance minimization (MPDM),
which ignores temporal dynamics and always assigns the closest available driver to a requesting
order~\cite{zhang2017taxi}. In Local
Policy Improvement~\cite{xu2018large}, handcrafted heuristics are combined with a machine learning method by summarizing
supply and demand patterns into a table and then using the learned patterns to account for future gains in the real-time
planner of the dispatching solution. A fully machine learning based approach to the dispatching problem is
presented in~\cite{wang2018deep}, where deep Q-learning is used for learning dispatching strategies
from the perspective of a single driver. We take these developments a step further and learn
fleet management and order dispatching strategies end-to-end using reinforcement learning.

Another thread of related work comes from the multi-agent reinforcement learning literature.
Specifically, our single-driver training approach is analogous to the ``independent Q-learning''
training approach~\cite{claus1998dynamics}.
Independent Q-learning has been leveraged successfully in a number of problems
including resource allocation problem for elevator control~\cite{crites1998elevator},
traffic signal control~\cite{eltantawy2010agent},
and fleet management~\cite{oda2018movi}.
Multi-agent reinforcement learning has also been investigated for order dispatching
in~\cite{li2019efficient}, where a grid-based algorithm is used.
Instead of operating on a grid, our method is based on continuous coordinates, which makes it more ready for deployment in the real-world.


\Skip{
\begin{enumerate}

    \item Dispatching
    \begin{enumerate}
        \item~\cite{zhang2017taxi} for KM
        \item~\cite{xu2018large} for tabular TD(0) + KM == LPI
        \item~\cite{wang2018deep} for DQN on single-driver scenario
        \item~\cite{oda2018movi} DQN for fleet management (movi)
    \end{enumerate}
    \item VRP and other combinatorial optimization papers
    \begin{enumerate}
        \item~\cite{vinyals2015pointer}, Proposes using an attention mechanism for solving combinatorial optimization problems. The network chooses which input token to emit next to the output.
        \item~\cite{kool2019attention}, Introduce GCNs instead of RNNs to the problem to use a stronger prior.
        \item~\cite{bengio2018machine}, A recent overview of ML approaches to combinatorial optimization problems.
    \end{enumerate}
    \item Meta-controller
    \begin{enumerate}
        \item Many StarCraft RL papers use meta-controllers to control a multitude of identical agents, kind of similarly to what our MDDQN does. Maybe check those for references for meta-controllers.
    \end{enumerate}
    \item Multiagent reinforcement learning \risto{put a lot less weight on this. Similar to treatment as in the AAAI submission would probably make sense}
    \begin{enumerate}
        \item~\cite{lanctot2017unified}
        \item~\cite{lowe2017multi} presents some challenges in MARL and proposes multi-agent adaptation of actor-critic methods
        \item~\cite{foerster2018learning}
        \item~\cite{pmlr-v70-foerster17b}
        \item~\cite{pmlr-v70-omidshafiei17a}
        \item~\cite{crandall2011learning} emergent communication. Maybe this approach deserves a short mention somewhere in the paper
        \item~\cite{sandholm1996multiagent} MARL for iterated prisoner's dilemma. An early work for applying PG to multi-agent problems.
    \end{enumerate}
\end{enumerate}
}

%% file: text/method.tex
\section{Method}
\label{sec:method}


\subsection{MDVDRP as a Reinforcement Learning Environment}
\label{subsec:mdvdrp}
The experiments presented in this paper are conducted in a ride-sharing simulation environment.
The environment is designed to capture the dynamic supply and demand situation
in ride-sharing platforms.
The environment represents customers as orders, which start and end in some
coordinates.
Assigning a driver to the order immediately yields a reward proportional to the price of the order.
Drivers are represented as points in the 2-D space and they can move by serving
orders or repositioning without an order.
The drivers and orders are generated following
a Poisson process with parameters depending on each scenario we consider.

Decision points are triggered whenever a driver becomes active in the system, finishes an order
or finishes a repositioning action. Variable amounts
of time may have passed between the decision points when the simulator polls the policy for actions.
The simulation scenarios considered in the experiments mostly depict the situation
where there are more customers demanding rides than there are drivers to serve them.
In this setting, it is natural to assume that only a single driver is available
for actions at any decision point.
This assumption simplifies the design of the policies by removing the need to choose
which driver to assign an order to.
To prevent multiple drivers polling for actions at exactly the same moment, we add random noise
with small variance to the duration of the reposition actions.
The actions available to the RL agent at each decision point are with respect
to the driver that is currently available in the environment.
These actions include assigning the driver to nearby orders as well
as actions to reposition the driver to
other areas.

Assigning a driver to an order removes the order from the system and makes the driver unavailable for
instructions until the order has been completed.
At order completion the driver is relocated to the order destination and made available for assignments.
Orders have a limited time window within which they are valid.
After the validity window has passed, the orders will be removed from the system.

The policies may choose to move the drivers to different directions for a fixed amount of time
by selecting reposition actions. The number
of available reposition actions includes actions that move the driver into one of the eight cardinal
directions for a fixed period of time and a stationary action.

The observation of the agents consists of the environment time, the driver for which action is currently
being selected and all drivers and orders currently in the simulation.
At time $t$ there is a collection of orders $o_t^i \in \mathcal{O}_t$, drivers $d^j_t \in \mathcal{D}^t$,
with exactly one {\em available driver} $d^{selected}_t$.
The state is given to the neural network as $s_t = (\mathcal{O}_t, \mathcal{D}_t, d_t^{selected}, t)$.
The orders are presented as 6 dimensional vectors consisting of the starting and ending coordinates, price, and time waiting. Time waiting is the difference between the
current time and the creation time. A driver is represented by a 6 dimensional
vector: the coordinates of the driver location, $x$ and $y$ components of its reposition direction, time to order completion, and time to reposition completion.
If the driver is serving an order, its location is set to the ending location of the order it is servicing.
If the driver is repositioning the
driver location is updated at each timestep, the reposition direction shows which way the driver will move during
the next timestep.

To limit the number of orders considered by the policy at each timestep, we impose a {\em broadcasting radius}
$d_{bcast}$ on the order assignment.
This means that drivers may be only paired with orders if they are within $d_{bcast}$ units of the
driver. Otherwise, the driver may only take a repositioning action. The repositioning actions
are not available to the drivers when there are orders within broadcasting radius of them.

\subsection{Reward Settings}
\label{subsec:reward_settings}

\begin{figure}
\centering
\includegraphics[width=0.47\textwidth]{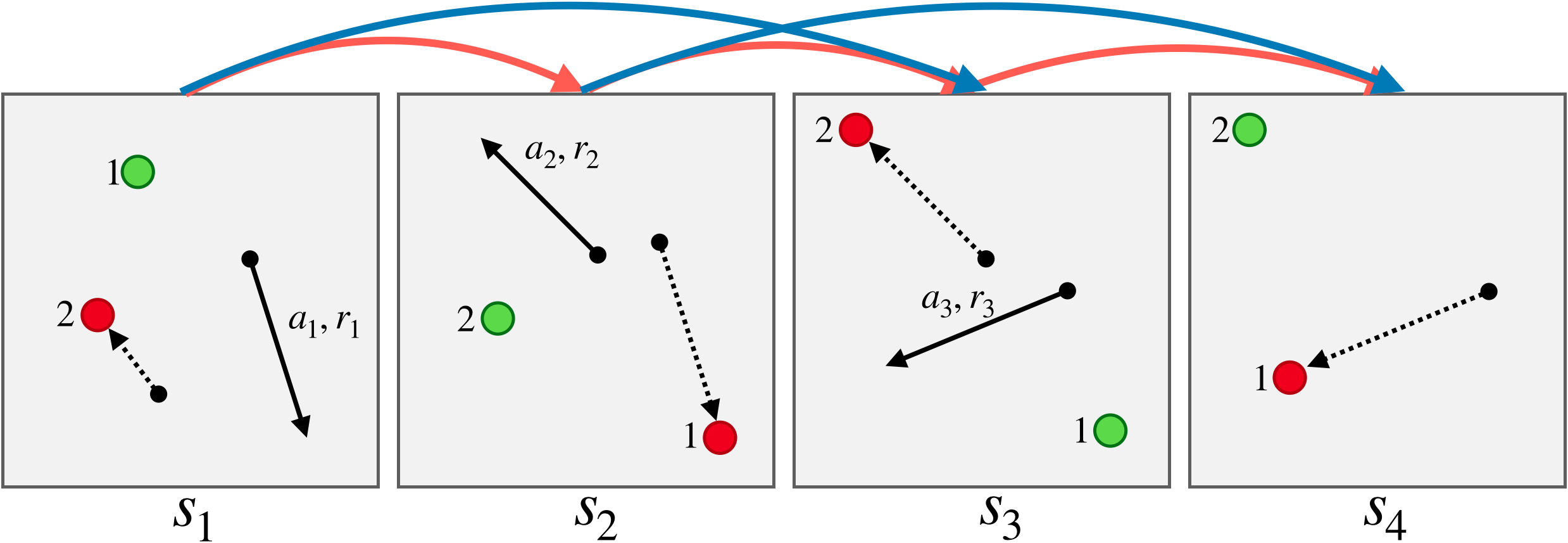}
\caption{The above image shows a trajectory with four timesteps in an MDVDRP instance with two drivers.
The currently available driver is green, dispatched driver is red, and the order that the available
driver accepts at time $t$ is $a_t$ and has price $r_t$. The accepted order at time $t$ is labeled
by its action name and price, $(a_t, r_t)$ and travels from the solid black dot to the terminal arrow.
Driver-centric transitions are indicated by blue arrows above state, e.g. transition
$(s_1, a_1, r_1, s_3)$, which is driver-centric with respect to driver 1. System-centric
transitions are indicated by red arrows e.g. transition$(s_1, a_1, r_1, s_2)$, which transitions
from a state where driver 1 is available to a state where driver 2 is available.}
\label{fig:transitions}
\end{figure}

The objective of the algorithms in the MDVDRP problem is to maximize the cumulative reward defined by the environment.
The environment rewards the agent for each assigned order with a reward the size of the order price.
Reposition actions yield no reward.
We consider two alternative reward specifications corresponding to \textit{driver-centric} and \textit{system-centric}
perspectives. An visual overview of these concepts is presented in \myfig{fig:transitions}

In the driver-centric approach, we consider the MDVDRP a reinforcement learning problem from the perspective of
the individual driver. In this setting, each driver is maximizing their own expected revenue and there are no
incentives for co-operation.
The trajectories taken by each driver in the environment are collected separately and the discounted returns
are computed on the individual trajectories. The individual trajectories consist of timesteps that are consecutive
from the perspective of the driver but not necessarily from the perspective of the environment as other drivers
may have taken actions between the actions of any single driver.
\Skip{
In more concrete terms, when a driver $i$
takes an action at timestep $t_1$, the agent immediately receives a reward $r_1$ for the action and the simulation
progresses until the next decision point at $t_2$. At $t_2$ the driver $j$ becomes available. Depending on the
state of the simulation, the drivers $i$ and $j$ may or may not be the same driver.
At some point in the simulation, on timestep $t_n$, the driver $i$ will be active again.
The driver-centric returns are computed by considering the timesteps $t_1$ and $t_n$ specific to the driver
$i$ as consecutive. The transitions between $t_1$ and $t_n$ will be accounted for on the trajectories
of other drivers.
}

From the point of view of the ride-sharing platform, the dispatching and repositioning problem is not about
individual drivers maximizing their own revenue but rather about the platform maximizing the combined
revenue across all drivers. Therefore, it is important to consider optimizing the policies from
the perspective of the whole system.
In this system-centric approach, the expected cumulative reward across all drivers is being maximized.
This leads to the trajectories experienced by the policy consisting of
timesteps that are consecutive from the perspective of the environment. For example, if the driver $i$
acts on the timestep $t_1$ receiving reward $r_1$ and driver $j$ takes an action on the timestep $t_2$,
the training algorithm will consider the timesteps $t_1$ and $t_2$ consecutive.

\subsection{Neural Network Architecture}

\begin{figure}
    \centering
    \includegraphics[width=0.4\textwidth]{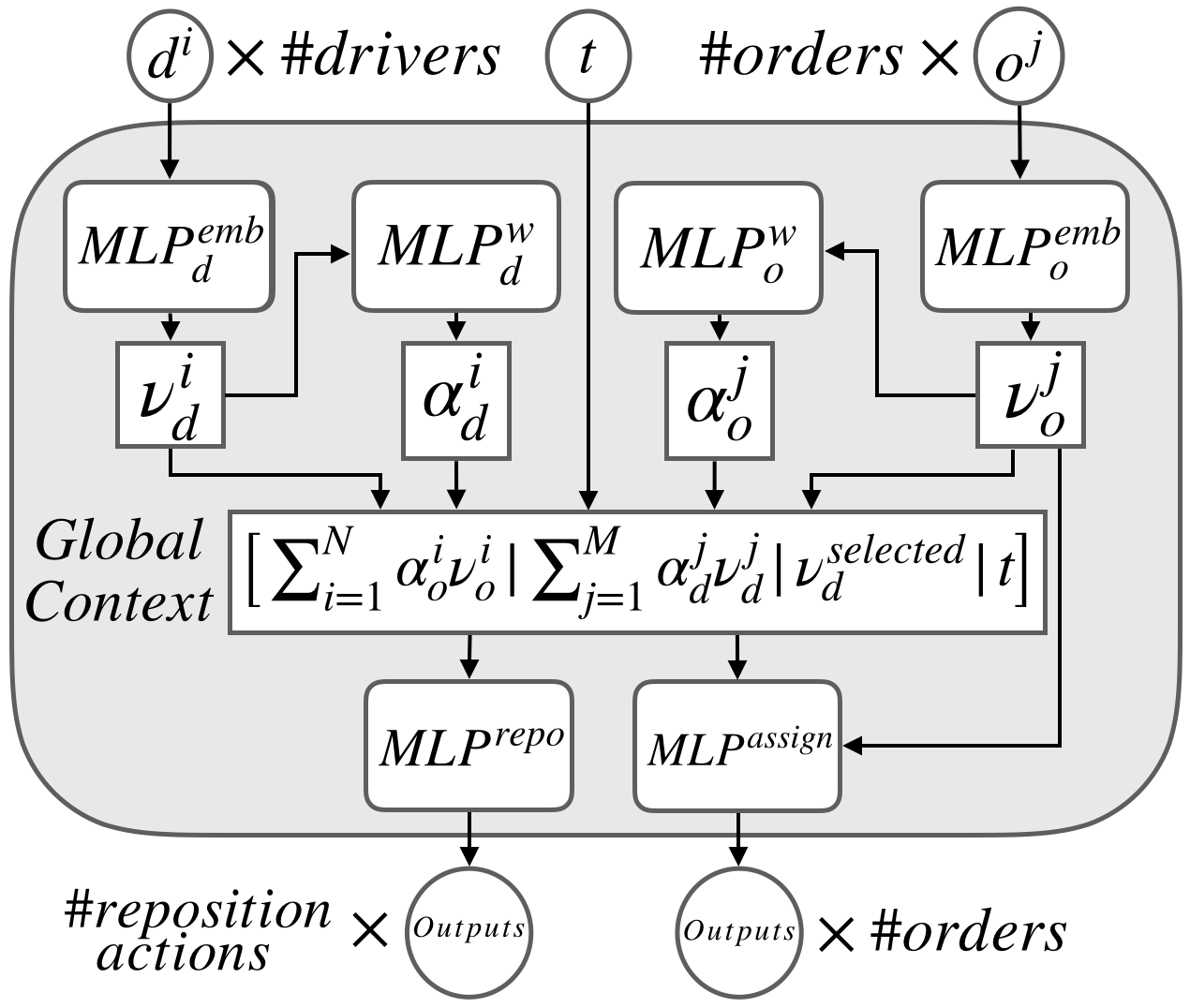}
    \caption{Network Architecture. Driver and order embeddings $\nu_d^i$, $\nu_o^j$
    and their corresponding weights $\alpha_d^i$, $\alpha_o^j$ are computed by $MLP^{emb}$ and $MLP^{w}$ respectively. A fixed sized global representation of the orders and drivers is computed by concatenating the sums of the weighted embeddings with the selected driver embedding and the environment time. 
    The fixed number of outputs for the reposition actions are computed by $MLP^{repo}$ using the global context as an input.
    $MLP^{assign}$ computes one output for each order using the global context and each order embedding as an input.}
    \label{fig:network_architecture}
\end{figure}

We propose a neural network architecture for RL in environments with variable sized
observation and action spaces.
An overview of the proposed network
architecture is presented in \myfig{fig:network_architecture}.
A learned pooling mechanism
allows the network to compute a fixed-sized global representation of the inputs, which
enables relating the features of each individual input to the global state.
In the environments we consider, the number of actions depends on the number of orders in the observation as described in~\ref{subsec:mdvdrp}.
We compute network outputs, one for each action, in a manner similar to the weight computation in the attention mechanism~\cite{mnih2014recurrent}.

The network first computes order embeddings $\nu_o^i$, order pooling weights $\alpha_o^i$, driver embeddings $\nu_d^j$, and driver pooling weights $\alpha_d^j$.
The embeddings are length 128 vectors computed by $MLP_{d}^{emb}$ and $MLP_{o}^{emb}$,
both of which have one hidden layer of size 128 and ReLU activations.
The scalar pooling weights $\alpha$ are computed from the embeddings by $MLP_{d}^{w}$ and $MLP_{o}^{w}$.
Both have hidden layer size 128 and tanh activation, and sigmoid output activation.
The global context vector is computed as $\big[\sum_{i=1}^{N} \alpha_o^i \nu_o^i|\sum_{j=1}^{M} \alpha_d^j \nu_d^j|\nu_d^{selected} | t\big]$, where $[a|b]$ denotes concatenation and $t$ denotes time.

$MLP^{assign}$, which has one hidden layer of size 64 and ReLU activation, outputs values for the assignment actions.
$MLP^{assign}$ takes each order embedding concatenated with the global context separately as input and produces a scalar output for each input.
The repositioning actions are computed by $MLP^{repo}$ with the same architecture as $MLP^{assign}$.
$MLP^{repo}$ has one output for each reposition action and uses the global context as its input.

\subsection{Training Algorithms}
\label{subsec:training_algorithm}
We train the proposed network architecture on the MDVDRP using two modern reinforcement learning algorithms: Deep Q-Networks (DQN)~\cite{mnih2015human} and Proximal Policy Optimization (PPO)~\cite{schulman2017proximal}. We chose these two algorithms because they represent widely used, modern methods in off-policy and on-policy reinforcement learning and many of the recent successes in deep reinforcement learning have been achieved using one of the two algorithms. A short description of the key ideas of the algorithms is given in the following. For more details we refer the reader to the corresponding papers.

DQN is a value-based, off-policy reinforcement learning algorithm. The Q-network estimates state-action values meaning it gives numeric estimates to the expected future return for each action available in the current state. On each iteration $i$, it optimizes the objective function
\begin{multline}
        \mathcal{L}(\theta_i) = \E_{(s, a, r, s')} \big[ (r + \gamma\argmax_{a'} Q(s', a'; \theta^{-}_i) \\
    - Q(s, a; \theta_i))^{2} \big]
\end{multline}
where $(s, a, r, s')$ are states, actions, rewards and next states sampled from the environment, $\gamma$ is a discount factor, $Q(s, a; \theta_i)$ is a neural network parameterized by $\theta_i$ approximating the action-value function, $\theta^-_i$ are the target network parameters and $\theta_i$ are the learning network parameters. The DQN algorithm collects experience from the environment, stores the experience in a replay buffer and updates the learning network parameters by performing stochastic gradient descent on the loss function. The target network is used to compute an action-value estimate in the next state $s'$.
In order to improve the stability of the learning network update, the target network parameters are copied from the learning network every couple of hundred update steps.
Actions are selected by taking the $\argmax$ of the Q-network output, except with probability $\epsilon$ a random action is chosen in order to make the algorithm explore. When evaluating the policies exploration is no longer needed and the epsilon can be set to a zero.

PPO is an on-policy actor-critic reinforcement learning algorithm~\cite{schulman2017proximal}.
Unlike DQN, PPO directly optimizes the policy using the policy gradient method.
In our case, the parameters of the policy are estimated by a neural network.
During training time the actions are chosen by sampling from the categorical distribution parameterized by the policy.
At evaluation time, we select the actions with the maximum probability.
The policy gradient objective is given by
\begin{equation}
    \nabla_{\theta}J(\pi_{\theta}) = \E_{\tau \sim \pi_{\theta}}
    \big[\sum_{t=0}^{T} \nabla_{\theta}\log \pi_{\theta}(a_t|s_t)A(\tau)\big]
\end{equation}
where $\pi_{\theta}(a_t|s_t)$ is the action distribution defined by the policy, $\theta$ are the parameters of the policy, $\tau = (s_0, a_0, r_0, ... s_T, a_T, r_T)$ are trajectories sampled from the environment, and $A(\tau)$ is the advantage function.
We estimate the advantage function using the generalized advantage estimation method~\cite{schulman2015high}, which requires requires computing a state-value function estimate called the critic.
The critic is learned alongside the policy using the same data.
On-policy reinforcement learning algorithms require new samples for each update they compute, otherwise they risk deviating too far from the previous policy, which may harm the policy performance.
Since sample collection can be expensive in reinforcement learning, updating on each batch of data multiple times would be beneficial.
In order to enable multiple updates on the same data, different variants of the PPO algorithm either clip or penalize the objective in order to prevent the policy from diverging too far from the previous one.
We use the clipping variant, which clips the objective based on the probability ratio between the current and the previous policy to remove the incentive to diverge too far.

As the simulation runs in continuous time, the time between consecutive timesteps may vary. Both algorithms use
a discounting factor $\gamma$ to compute the discounted return, which accounts for the future rewards.
In continuous time settings, the discounting factor for each timestep is $\gamma^{t' - t}$,
where $t$ is the time at the current observation and $t'$ in the next.

We found empirically that training using the system-centric rewards is more challenging than using the
driver-centric rewards. We found the use of n-step Q-learning~\cite{mnih2016asynchronous} helpful for
stabilizing the training of DQN when training on system-centric rewards. In all of our DQN experiments
with system-centric reward we use 20-step Q-learning.

%% file: text/experiments.tex
\section{Experiments}
\label{sec:experiments}

\begin{figure}[t]
\centering
\begin{tabular}{cc}
     \includegraphics[width=.22\textwidth]{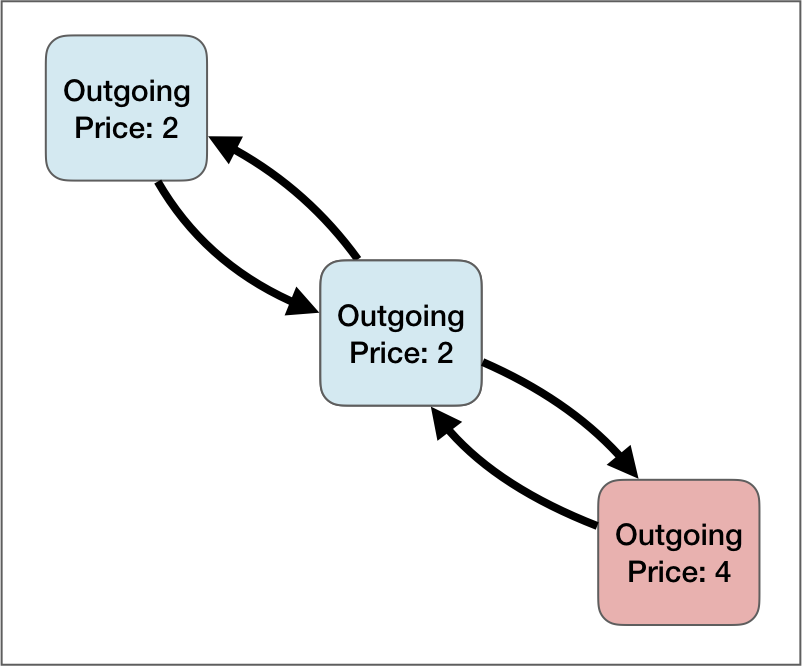} &
     \includegraphics[width=0.22\textwidth]{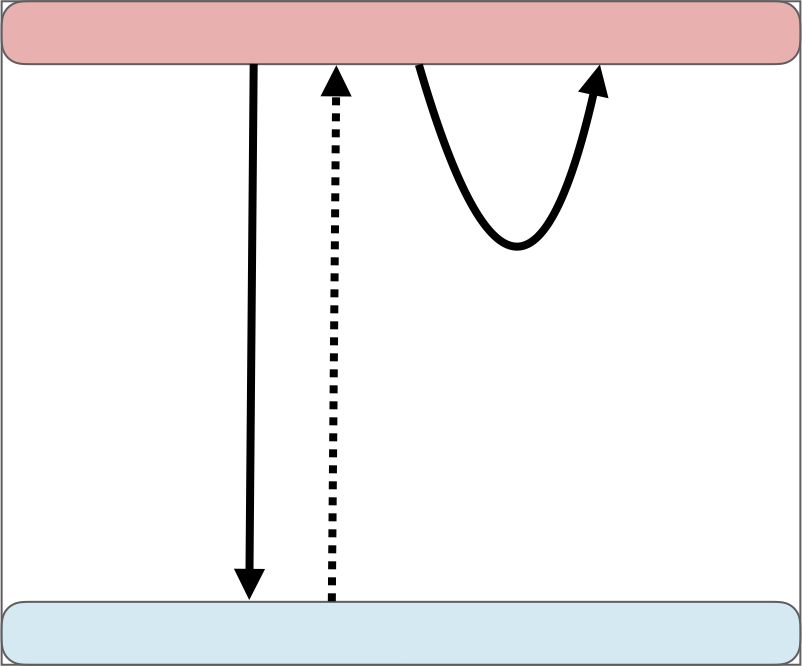} \\
     \small{(a) Regional Domain} & \small{(b) Hot-Cold Domain} \\
    \includegraphics[width=.22\textwidth]{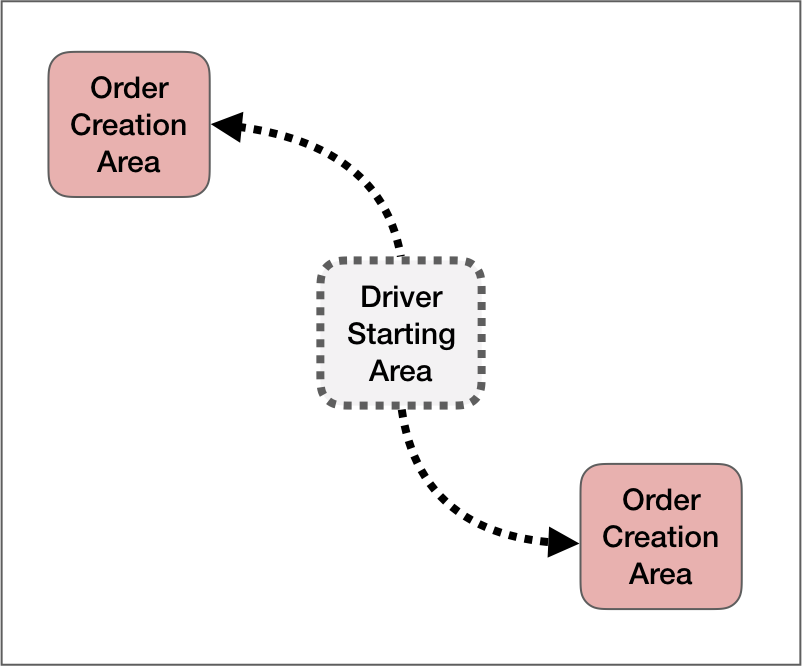} &
    \includegraphics[trim=0 50 0 75, clip, width=0.22\textwidth]{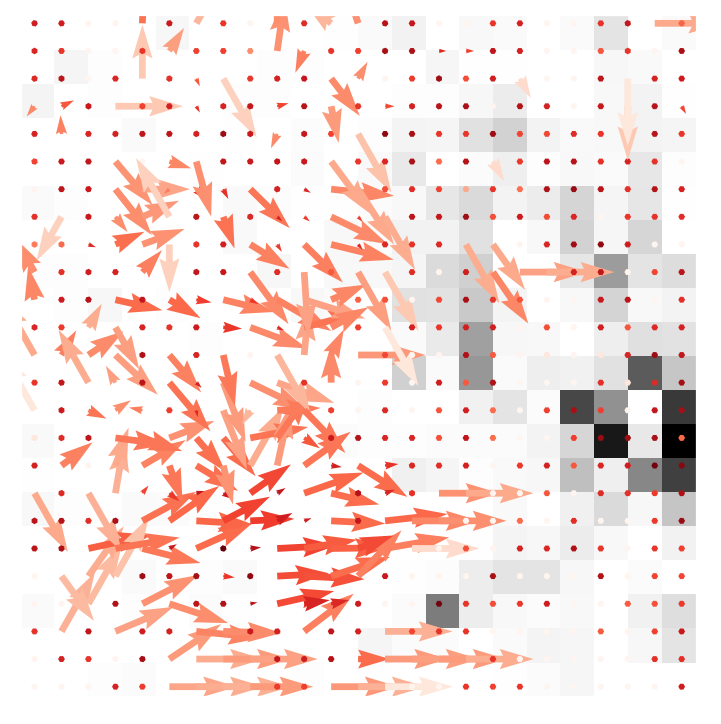} \\
    \small{(c) Distribute Domain} & \small{(d) Historical Orders Domain}
\end{tabular}
\caption{Conceptual drawings of the illustrative domains considered in the experiments. The colored regions represent areas where orders are generated. The solid arrows depict order starting and ending locations and the dotted paths depict paths the drivers need to reposition along in order to get to the orders. Best viewed in color.
(a) \textbf{Regional domain} tests whether the policies can learn to exploit price differences between otherwise similar orders. Each square is labeled with its outgoing order value.
(b) \textbf{Hot-Cold Domain} seeks to evaluate the algorithms ability to learn allocating drivers to orders which minimize driver idle time. All orders begin in the red bar, with their positions generated uniformly randomly. For the destination, a fair coin is flipped to decide whether the order ends in hot or cold, and then the exact position is sampled uniformly randomly in the designated region.
(c) \textbf{Distribute Domain} requires the policies to anticipate that the orders are going to be created in the two distant regions and allocate correct number of drivers to each.
Figure (d) represents an order generation pattern defined by the historical data and the corresponding reposition strategy learned by the policy for an hour long time window in \textbf{Historical Data Domain}. The graph is composed of 2d histogram depicting the order generation pattern and arrow plots depicting the average reposition movements in each histogram bin. The intensity of the arrow color corresponds to the number of drivers who have been in each cell during the time window.}
\label{fig:illustrative_envs}
\end{figure}

\input{text/result_table.tex}
\input{text/distribute_table.tex}

We investigate the learning behavior of the proposed network architecture in ride-sharing
environments implemented using the simulator described in \ref{subsec:mdvdrp}.
We present results in five
environments to test and illustrate the dispatching and repositioning strategies learned the
proposed approach. The environments are illustrated in Figures \ref{fig:illustrative_envs},
\ref{fig:historical_order_histogram}.
The geography
of the environments is presented as a rectangle with the longer side length set to one.
The drivers move at speed 0.1 and the broadcasting radius is set to 0.3.

We compare the learned policies against two kinds of baselines: myopic revenue maximization (MRM)
and myopic pickup distance minimization (MPDM) \cite{zhang2017taxi}. MRM always assigns the
highest value order to the closest available driver. MPDM assigns orders to drivers in the
order of shortest distance first. Since the baselines do not account for repositioning, we test
three variants of the baselines with different repositioning heuristics. In the simple variants,
broadcasting distance is ignored and drivers can be assigned to orders across the region. The
drivers stay where they drop orders off until they are assigned to the next order. In
the \textit{random} variants of MRM and MPDM, if a driver has no orders the within broadcast
distance, then a repositioning action is selected randomly. The \textit{demand} variants of MRM
and MPDM apply a simple heuristic on repositioning, by moving the drivers towards the
nearest order when an order is available but not within broadcasting distance.
If there is no order available, the demand variants take random reposition actions.

We use the same neural network architecture for all variants of the learning algorithms,
except for the extra output layer implementing the critic for PPO, which does not
exist in the model used for DQN.
We set the discount factor $\gamma = 0.99$ for both
algorithms. For DQN we use batch size of $32$, replay buffer size $20000$ and update on every
timestep with learning rate $0.0001$. The exploration rate $\epsilon$ starts from $0.99$ and
we anneal it by $0.01$ on each episode to the final value $0.1$. We update the target network
every $100$ steps.
The PPO policy and value functions are updated $20$ times every epoch using $4000$ consecutive
timesteps as the training data.
We use generalized advantage estimation (GAE) \cite{schulman2015high} and set the GAE hyperparameter
$\lambda = 0.95$. We aim to keep
hyperparameters unchanged between the different environments but make slight changes for
Distribute Domain and the real data based domains. Those changes are explained in the relevant
sections. To handle the large numbers of drivers and orders in the Historical Statistics
domain, we use a parallel implementation of PPO and suitable hyperparameters, which are
explained in \ref{subsec:historical_stats}.

The results from the experiments can be found in tables \ref{table:results} and
\ref{table:distribute}. We present training curves for the Hot-Cold, Regional
and real data based environments in Figures \ref{fig:dqn_training_curves},
\ref{fig:ppo_training_curves} and \ref{fig:real_data_training_curves}.

\subsection{Regional Domain}
\label{subsec:regional_domain}
In our first experiment, we consider the Regional Domain, which illustrates how a simple price differential can be exploited
by learned policies but is missed by the myopic dispatching and repositioning approaches.
Intuitively, in this domain, a good policy dedicates enough drivers to fully serve the high-reward orders and serves the other orders with the remaining drivers.
Myopic policies, which ignore the long-term effects of their decisions, will fail to exploit the available extra reward from serving the high-reward orders.
The high-level concept of Regional Domain is presented in \myfig{fig:illustrative_envs}. In the Regional Domain, there are three regions: left, center, and
right. One quarter of all orders go from the center to the upper-left region, one quarter from center to bottom-right,
one quarter from upper-left to center, and one quarter from bottom-right to center. All orders yield a reward of 2
except those that go from right to center, which yield a reward of 4.

In the high demand version of the Regional Domain, the proposed method outperforms the baselines by over a 15\% margin between
the best performing baseline (MPDM-demand) and the best learned policy (Driver-Centric DQN).
To identify the source of the difference between the learned policies and the baselines, we look
at a policy trained with PPO on driver-centric rewards. The learned policy serves 70 times more orders from the high reward
region than the low reward region and achieves significantly higher score than the baselines, which suggests that it
has learned to exploit the higher priced orders to its advantage.

The low demand variant proves to be challenging for the
learning algorithms. Here the demand is low enough that over
commitment to the high reward area can become a problem as is the case for Driver-Centric PPO,
which ends up allocating 8 times more
idle drivers to the high reward area and as a consequence has $8\%$ lower ratio of total orders
served compared to MPDM-demand. This policy leads to weak performance, and is in fact outperformed
by the simple and demand variants of MPDM.

\subsection{Hot-Cold Domain}
\label{subsec:hot_cold_domain}
The Hot-Cold domain can be thought of as a ride-sharing scenario with a busy area of downtown
(``hot region'') surrounded by suburbs with less traffic (``cold region'').
In the Regional domain, the advantage of traveling to the region with higher prices is clear;
it is directly tied to the price of orders found in that region (4 vs. 2).
In the Hot-Cold domain, the agent must learn a more subtle advantage.
Order pickup locations are located uniformly along the top edge of the simulation area.
Half of the orders end uniformly along the bottom edge of the area and half end uniformly in the hot region.
Order price is given by the Euclidean distance from order pickup to order drop-off locations.
Despite orders to the cold region having higher price (since they are longer on average),
it is generally more advantageous for drivers to stay in the hot region, since there they can quickly pick
up new orders. In other words, the advantage is entirely {\em temporal}.
An illustration of the Hot-Cold domain is presented in \myfig{fig:illustrative_envs}.

The results in \mytable{table:results} suggest that learning to balance serving the orders to
hot and cold regions is straightforward for the learning algorithms and all learning
algorithms outperform all baselines in the high demand variant of the environment. As with the Regional domain,
the challenges of training DQN with system-centric rewards are apparent in the low demand Hot-Cold domain,
where it is outperformed by multiple baselines.

The margin in favor of the learned policies is narrower in the low demand variant.
In the low demand variant of the environment, the policies trained with driver centric rewards the ratio
of orders served is approximately double compared to the high demand variant (50\% vs 25\%).
The higher served ratio suggests that there is less freedom of choice between hot-to-hot and
hot-to-cold orders in the low demand regime, which in turn reduces the potential for
improvement over the naive policies.

In all of the experiment configurations for Regional domain and Hot-Cold domain, we found DQN to be more
sample efficient than PPO. DQN was trained for 1000 episodes and PPO often took two to three times
as many to achieve similar performance. Actor-critic methods are known to be less sample
efficient than value-based methods and another contributing factor may be the relatively low learning rate we used for PPO.

\subsection{Distribute Domain}
\label{subsec:distribute}
While Hot-Cold domain tests an important aspect of learning - namely, the ability of the agents to reposition
drivers to locations where they can pick up new orders, this repositioning behavior is quite simple in that
it is {\em uniform across drivers}. This means that each individual driver can always be repositioned in the
same manner (\ie "if in cold region, go to hot region"). In order to test whether our methods can learn
non-uniform repositioning behavior, we introduce a class of ``distribution environments'' where drivers must
be repositioned so as to match their spatial distribution with a fixed future order distribution. A Distribute
domain operates in two phases. In the first phase, the environment resets to a state with $k$ drivers and no
orders, so drivers may only reposition during this phase. In the second phase, $k$ orders appear according to
a fixed spatial distribution, and drivers can match to orders if they are within a given broadcast radius
$d_{bcast}$. The second phase only lasts long enough to allow drivers to reposition one more time before all
orders cancel and the environment is reset. Each order matching action receives $+1$ reward. Order
destinations are placed away from start locations so that each driver may only serve one order per episode.
As a result, the episodic return is proportional to the number of orders served, so we may interpret the
episode score as a measure of how well the agent arranges driver supply in phase 1 with order demand in
phase 2.

In our experiments, the distribution of orders always consists of two small patches in the top left and bottom right parts of the unit square, refer to \myfig{fig:illustrative_envs} for visualization. The order start locations are sampled uniformly within each patch. The total number of orders in each patch is fixed across episodes, and we denote it fractionally. An even order split between patches (\eg 10 orders in both patches) is denoted $50/50$. If 80 percent of orders are in the first patch and 20 percent are in the second patch, we denote it as $80/20$.

We found that training successful policies on the Distribute Domains requires more random exploration than any of the other environments. For DQN, we set the final $\epsilon=0.2$. For PPO, we follow an annealing scheme similar to DQN where we anneal the entropy coefficient from $0.7$ to $0.01$ over the first 2000 epochs.

Results for distribute domains with 20 drivers are presented in \mytable{table:distribute}.
We include the optimal served percentage (which is 100 \%) and the ``uniform optimal''
served percentage. This quantity reflects the maximum score one can obtain if the
repositioning behavior is uniform across drivers. The results between driver and system-centric
variants are mixed. While system-centric PPO achieves a slight
advantage over the driver-centric variant, the relationship is flipped for DQN.

\subsection{Historical Orders Domain}
\label{subsec:historical}
\begin{figure}
    \centering
    \begin{tabular}{cc}
         \includegraphics[width=0.22\textwidth]{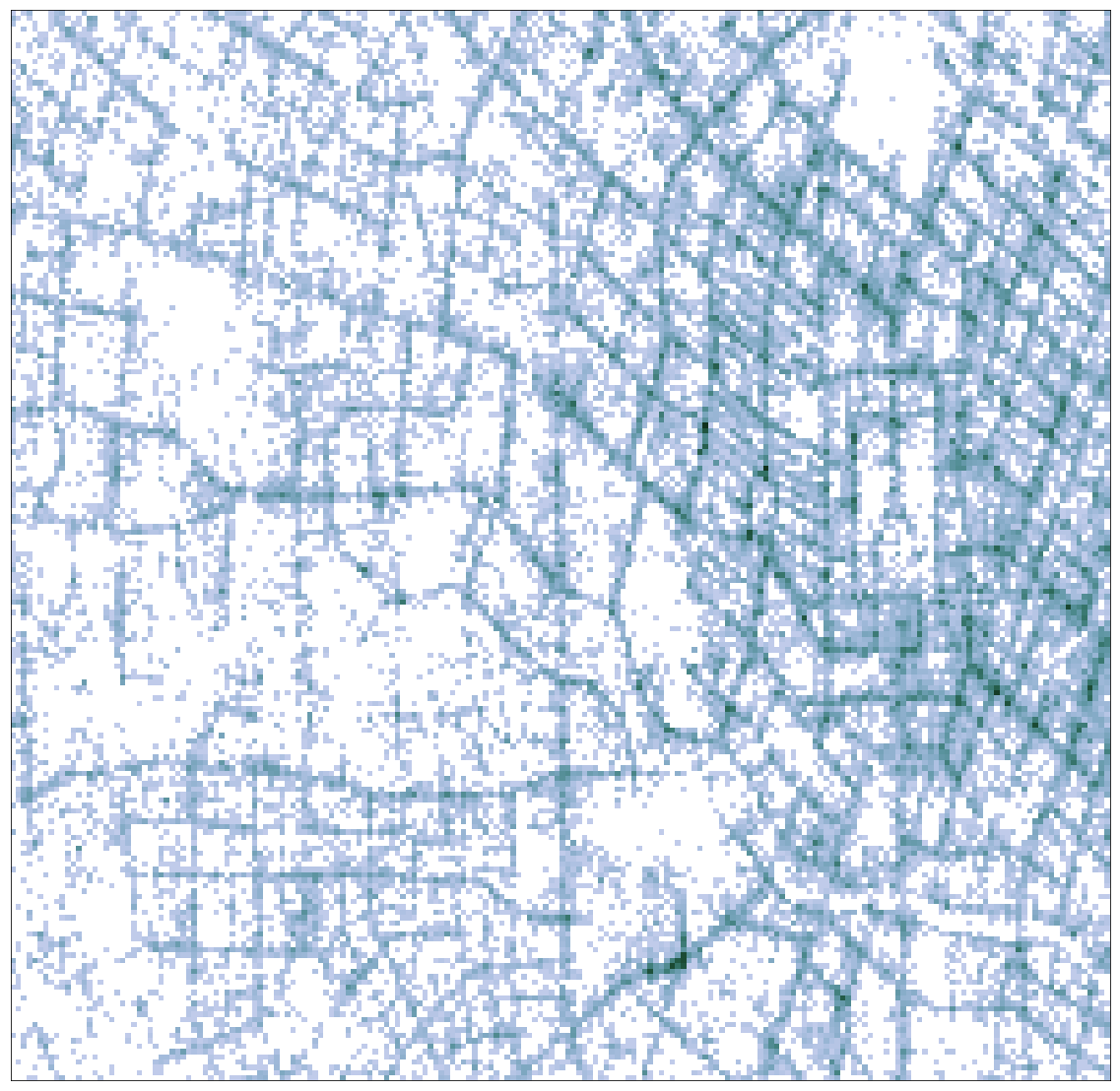} &
         \includegraphics[width=0.22\textwidth]{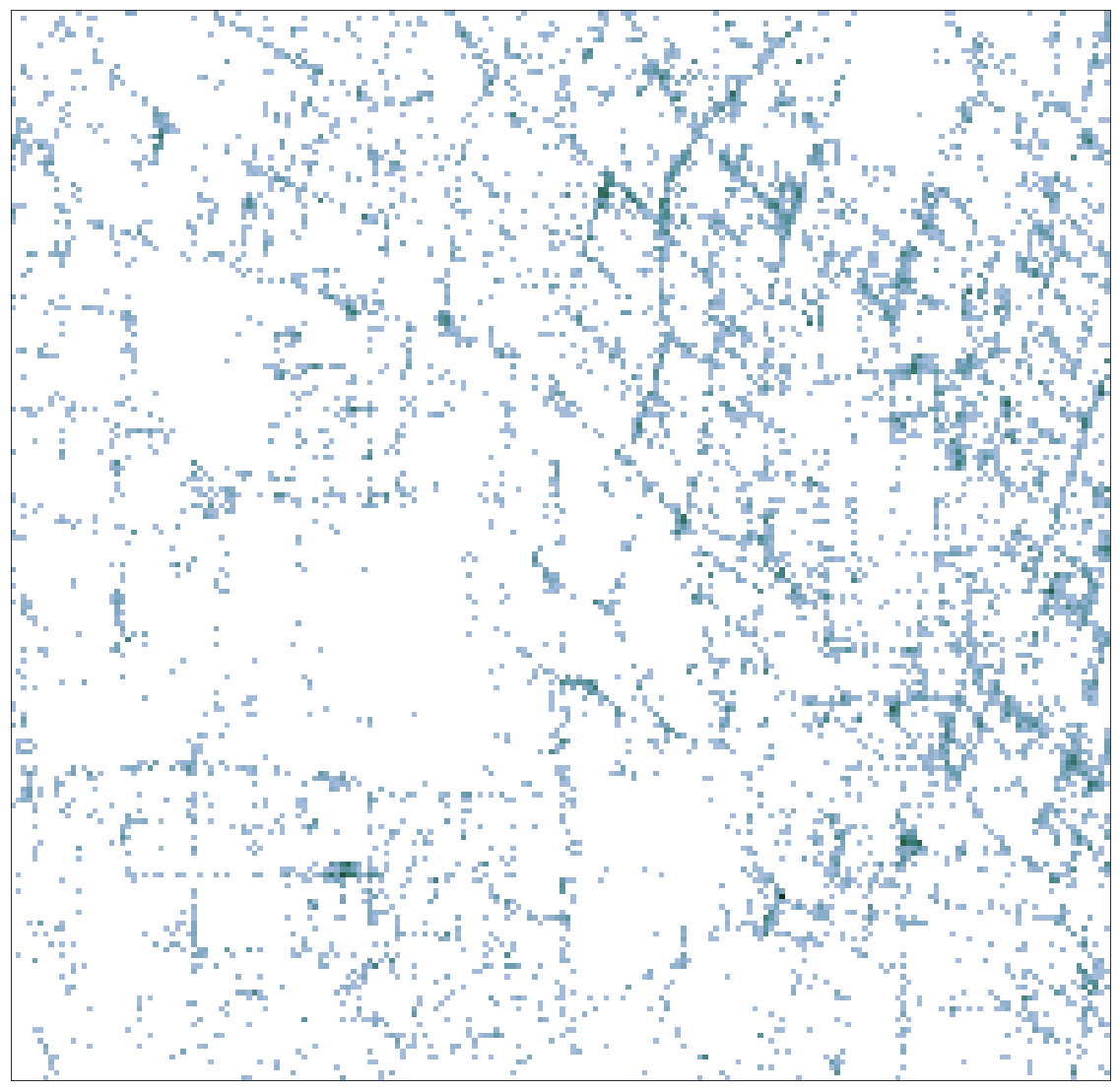} \\
         \textbf{(a) Active Hours} & \textbf{(b) Quiet Hours} \\
    \end{tabular}
    \caption{2d histograms depicting the order creation in the {\em Historical Data Domain}.
    Darker colors indicate more orders requested in the cell according to a power law
    coloring scheme. The histograms colors have been determined by normalizing the histograms separately for visual clarity.
    In \textbf{(a)} the histogram corresponds to the 6 most active hours on all of the 30 days
    during which the dataset has 294279 orders for the region under consideration.
    The 6 quietest hours are similarly displayed in \textbf{(b)}, which corresponds to
    21807 orders in the dataset.}
    \label{fig:historical_order_histogram}
\end{figure}

Solving the MDVDRP is motivated by the real-world need of ride-sharing platforms to improve
their marketplace for drivers and passengers.
To test our learning algorithms in a more realistic setting, we consider an environment where the order
generation scheme is based on historical order data from the GAIA dataset \cite{gaia2017}. The dataset provides gps records
for all orders served during a 30-day period in the city of Chengdu in China.
To limit computational demands, we choose a subset of consisting of $10\%$ of the orders in the dataset.
The orders in the subset all start and end in an area from the dataset that has approximately twenty
thousand orders per day.
We then create 30 {\em order generation schemes}, which correspond to the 30 days in the dataset.
Each episode in the environment corresponds to a randomly sampled day from the dataset. The
orders generated during the episode appear exactly in the coordinates and at the time defined by the data.
We use a fixed number of drivers (100), a 2 kilometre broadcast radius and a fixed speed (40 km/h).
Histograms of the order start locations in the area under consideration are presented in
\myfig{fig:historical_order_histogram}.

Similarly to the Hot-Cold domain, the reward for serving an order is the distance
from the order start location to the drop-off location. Following from the reward definition,
the total reward each driver receives is directly correlated with the total time
they spend serving orders and thus inversely correlated with the time spend between orders.

Driver-centric DQN and PPO are both capable of learning strong dispatching and repositioning policies,
which outperform all of the baselines by over 6\%. The learning curves for the experiments in this
domain are presented in \myfig{fig:real_data_training_curves}.
The historical data domain also demonstrates the importance of redistributing drivers during their downtime.
The worst performing algorithms on the historical data domain are the {\em simple variants} of MPDM and MRM,
which do not use reposition actions at all. In contrast, the learned policies actively redistribute
drivers back to the populated areas as shown in Figure \myfig{fig:illustrative_envs} (d).

Finally, we observe that
both DQN and PPO have trouble learning competitive policies on this larger scale problem when trained
on the system-centric rewards. We hypothesize that this is due to having more drivers complicating
the already challenging credit assignment problem. With more drivers simultaneously acting in the
environment, the time between decision points becomes smaller and the effective discount factor
approaches one due to the way it is computed. Training with a discount factor close
to one can make the training unstable. To mitigate this effect we use a smaller discount factor
$\gamma = 0.9$.

\subsection{Historical Statistics Domain}
\label{subsec:historical_stats}
Finally, we test our method in another MDVDRP instance based on the GAIA dataset.
In the Historical Statistics domain, both drivers and orders are generated from a Poisson process.
Having the number of drivers vary with the time of the day takes the simulation another step closer
to the realistic setting.

To build the simulator from this data, we first covered the city in a square 20 by 20 grid, and extracted
Poisson parameters $\kappa_{x,y,t}$ where $x$ is an order start tile, $y$ is an order end tile,
and $t$ is the time of day in hours. This results in $400 \times 400 \times 24 = 3.84$ million
parameters which we use specify an order generation process. The order generation parameters for the
most active and the most quiet hours in the day averaged over all 30 days.
The price of an order is determined by the trip distance.

We determine the driver generation parameters using a similar approach as we used for the order parameters.
A driver is counted to have become active when it serves its first order during the day. The drivers remain
active for six hours after they have been created. In the initial testing of the environment,
we found that using the raw driver generation parameters results in enough drivers being created
that even a randomly initialized policy is able to achieve above 99\% ratio of orders served.
The dataset only records orders which were served by drivers, so it seems natural that the
served ratio using the unscaled parameters would be high.
The approximations we make about the duration drivers remain active and the speed of the drivers
could also be contributing to the high served ratio.
In order to set up the environment such that there is room
for improvement, we scale the driver generation parameters down to 7\%.
The driver generation parameters are further scaled down by a factor of 0.5, which we also
apply to the order generation parameters, to lower the computational demands.

To handle the large scale of the domain with close to 100000 orders per day served by over a 1000
drivers, we use a parallelized implementation of PPO. Using parallel environments allows us
to sample shorter trajectories in each environment to achieve the same update batch size.
Sampling fewer consecutive transitions between each update speed up the training.
We use 10 parallel environments and correspondigly reduce the steps per epoch hyperparameter
to 400. We also raise the policy learning rate to $0.0005$ and value function learning
rate to $0.001$.

The training curves from the experiments are presented in \myfig{fig:real_data_training_curves}.
Training with either system and driver-centric approaches results in policies that clearly outperform the
baselines. The best performing learned policy is the driver-centric PPO, which achieves the evaluation score
of $286135 \pm 1089$. For system-centric PPO the evaluation score is $ 261059 \pm 184 $ and the best performing
baseline, MRM-random, scores $ 252656 \pm 280 $. The driver-centric approach outperforms the system-centric
by a margin of close to $10\%$.
It is evident from the evaluation score and the learning curves that learning with the system-centric rewards
is challenging, possibly even more so as in the Historical Orders domain.

%% file: text/result_table.tex
\Skip{

\begin{table*}
\begin{center}
    \caption{\textbf{Experimental Results.} The algorithms were evaluated multiple times during the training at fixed intervals and the results of the best evaluation step across all random seeds is reported $\pm$ standard error. The reported values are averages over 5 episodes for Hot Cold and Regional domains and over 20 episodes for Historical Orders and Historical Statistics domains. All the scores that are within standard error of the best score are bolded in the table.}
    \resizebox{\textwidth}{!} {
        \begin{tabular}{||c | c c | c c | c | c ||} 
            \hline
            \multicolumn{1}{|c}{} & \multicolumn{2}{|c|}{Hot Cold} & \multicolumn{2}{|c|}{Regional} & \multicolumn{1}{|c|}{Historical Orders} & \multicolumn{1}{c|}{Historical Statistics} \\
            \hline Algorithm & High Demand & Low Demand & High Demand & Low Demand &  &\\ [0.5ex] 
            \hline\hline
            MRM-simple  & $ 5359 \pm 8$   & $ 5189 \pm 14 $ & $ 3597 \pm 6$   & $ 2964 \pm 22$  & $  40001 \pm 128 $ &  $ 185690 \pm 261 $\\
            \hline
            MPDM-simple & $ 5917 \pm 6$   & $ 5713 \pm 20 $ & $ 4258 \pm 8$   & $3328 \pm 28 $  & $  37960 \pm 127 $ &  $ 156539 \pm 167 $\\
            \hline
            MRM-random  & $ 797  \pm 30$   & $ 949 \pm 41 $ & $ 2150 \pm 30$  & $ 3039 \pm 30$  & $  49563 \pm 309 $ & $ 252656 \pm 280 $ \\
            \hline
            MPDM-random & $ 1006 \pm 35$  & $ 1004 \pm 50 $ & $ 4203 \pm 19$  & $ 3103 \pm 33$  & $  46763 \pm 241 $ & $ 241401 \pm 357 $ \\
            \hline
            MRM-demand  & $ 5351 \pm 9$   & $ 5449 \pm  9$  & $ 2161 \pm 28$  & $ 3262 \pm 16$  & $  48805 \pm 463 $ & $ 252268 \pm 279 $  \\
            \hline
            MPDM-demand & $ 5883 \pm 17$  & $ 5658 \pm 15 $ & $ 4252 \pm  8$  & $ \mathbf{3343 \pm 16}$  & $  46635 \pm 539 $ & $ 241381 \pm 216 $ \\
            \hline
            PPO System-Centric      & $ \mathbf{7954 \pm 17} $  & $ 5801 \pm 29 $ & $ 4744 \pm 2$   & $ \mathbf{3372 \pm 33}$  & $  50094 \pm 162 $ & $ 261059 \pm 184 $   \\
            \hline
            DQN System-Centric     & $ 6323 \pm 168$ & $ 5278 \pm 77 $ & $ 4735 \pm 12$  & $ 2831 \pm 104$ & $  48532 \pm 71 $ & N/A \\
            \hline
            PPO Driver-Centric      & $ 7861 \pm 23$  & $ 5767 \pm 10 $ & $ 4888 \pm 2$   & $ 3208 \pm 16$  & $  53029 \pm 45 $ & $ \mathbf{ 286135 \pm 1089 }$ \\
            \hline
            DQN Driver-Centric      & $ 7883 \pm 3$  & $ \mathbf{5855 \pm 8} $  & $ \mathbf{5006 \pm 15}$   & $ \mathbf{3349 \pm 6} $  & $  \mathbf{53255 \pm 130}$ & N/A \\
        \hline 
        \end{tabular}
        \label{table:results}
    }
    
        \end{center}
\end{table*}

}

\begin{table*}
\begin{center}
    \caption{\textbf{Experimental Results.} The algorithms were evaluated multiple times during the training at fixed intervals
    and the results of the best evaluation step across all random seeds is reported $\pm$ standard error. The reported
    values are averages over 5 episodes for Hot Cold and Regional domains and over 20 episodes for Historical Orders domain.
    All the scores that are within standard error of the best score are bolded in the table.}
    \resizebox{0.9\textwidth}{!} {
        \begin{tabular}{||c | c c | c c | c  ||} 
            \hline
            \multicolumn{1}{|c}{} & \multicolumn{2}{|c|}{Hot Cold} & \multicolumn{2}{|c|}{Regional} & \multicolumn{1}{c|}{Historical Orders}  \\
            \hline Algorithm & High Demand & Low Demand & High Demand & Low Demand & \\ [0.5ex] 
            \hline\hline
            MRM-simple  & $ 5359 \pm 8$   & $ 5189 \pm 14 $ & $ 3597 \pm 6$   & $ 2964 \pm 22$  & $  40001 \pm 128 $ \\
            \hline
            MPDM-simple & $ 5917 \pm 6$   & $ 5713 \pm 20 $ & $ 4258 \pm 8$   & $3328 \pm 28 $  & $  37960 \pm 127 $ \\
            \hline
            MRM-random  & $ 797  \pm 30$   & $ 949 \pm 41 $ & $ 2150 \pm 30$  & $ 3039 \pm 30$  & $  49563 \pm 309 $ \\
            \hline
            MPDM-random & $ 1006 \pm 35$  & $ 1004 \pm 50 $ & $ 4203 \pm 19$  & $ 3103 \pm 33$  & $  46763 \pm 241 $ \\
            \hline
            MRM-demand  & $ 5351 \pm 9$   & $ 5449 \pm  9$  & $ 2161 \pm 28$  & $ 3262 \pm 16$  & $  48805 \pm 463 $ \\
            \hline
            MPDM-demand & $ 5883 \pm 17$  & $ 5658 \pm 15 $ & $ 4252 \pm  8$  & $ \mathbf{3343 \pm 16}$  & $  46635 \pm 539 $ \\
            \hline
            PPO System-Centric      & $ \mathbf{7954 \pm 17} $  & $ 5801 \pm 29 $ & $ 4744 \pm 2$   & $ \mathbf{3372 \pm 33}$  & $  50094 \pm 162 $ \\
            \hline
            DQN System-Centric     & $ 6323 \pm 168$ & $ 5278 \pm 77 $ & $ 4735 \pm 12$  & $ 2831 \pm 104$ & $  48532 \pm 71 $  \\
            \hline
            PPO Driver-Centric      & $ 7861 \pm 23$  & $ 5767 \pm 10 $ & $ 4888 \pm 2$   & $ 3208 \pm 16$  & $  53029 \pm 45 $ \\
            \hline
            DQN Driver-Centric      & $ 7883 \pm 3$  & $ \mathbf{5855 \pm 8} $  & $ \mathbf{5006 \pm 15}$   & $ \mathbf{3349 \pm 6} $  & $  \mathbf{53255 \pm 130}$ \\
        \hline 
        \end{tabular}
        \label{table:results}
    }
    
        \end{center}
\end{table*}

%% file: text/distribute_table.tex
\begin{table}
\begin{center}
    \caption{Distribute Domain with 20 Drivers}
    \resizebox{\columnwidth}{!}{
        \begin{tabular}{||c | c | c ||}
            \hline Algorithm & 50/50 Served \% & 80/20 Served \% \\ [0.5ex] 
            \hline
            Optimal & $100 \% $ & $ 100 \% $ \\ 
            \hline
            Uniform Optimal & $50 \% $ & $80 \% $ \\
            \hline
            PPO System-Centric & $100 \pm 0.0 \% $ & $93 \pm .54 \%$  \\
            \hline
            DQN System-Centric & $95 \pm .11 \%$ & $80 \pm 3.42 \% $ \\
            \hline
            PPO Driver-Centric & $96 \pm .13 \% $ & $92 \pm .72 \%$  \\
            \hline
            DQN Driver-Centric & $96 \pm .13 \% $ & $92 \pm .72 \%$  \\
        \hline 
        \end{tabular}
    }
    
\label{table:distribute}
\end{center}
\end{table}

%% file: text/conclusion.tex
\section{Conclusion}
\label{sec:conclusion}
We performed a detailed empirical study of reinforcement learning approaches to multi-driver
vehicle dispatching and repositioning problems.
We studied driver-centric and system-centric reward definitions and trained policies using
DQN and PPO algorithms. We found DQN to be more sample efficient than PPO. On the other hand
PPO was better able to learn on system-centric rewards.
Central to all of our approaches was the network architecture we
presented, which leverages a global representation of state processed using attention mechanisms.
We found that, while one can construct environments
where the system-centric approach is superior, typically driver-centric is better or at least competitive
with the system-centric approach.
Furthermore we applied these methods to environments built using real dispatching data,
and found that driver-centric approach is able to consistently beat myopic dispatching and
repositioning strategies.


%% file: text/appendix.tex
\begin{figure*}
    \centering
    \begin{tabular}{cccc}
        Hot-Cold High Demand & Hot-Cold Low Demand & Regional High Demand & Regional Low Demand \\
        \includegraphics[width=0.22\textwidth,trim={1.8cm 4.0cm 5.0cm 4.0cm},clip]{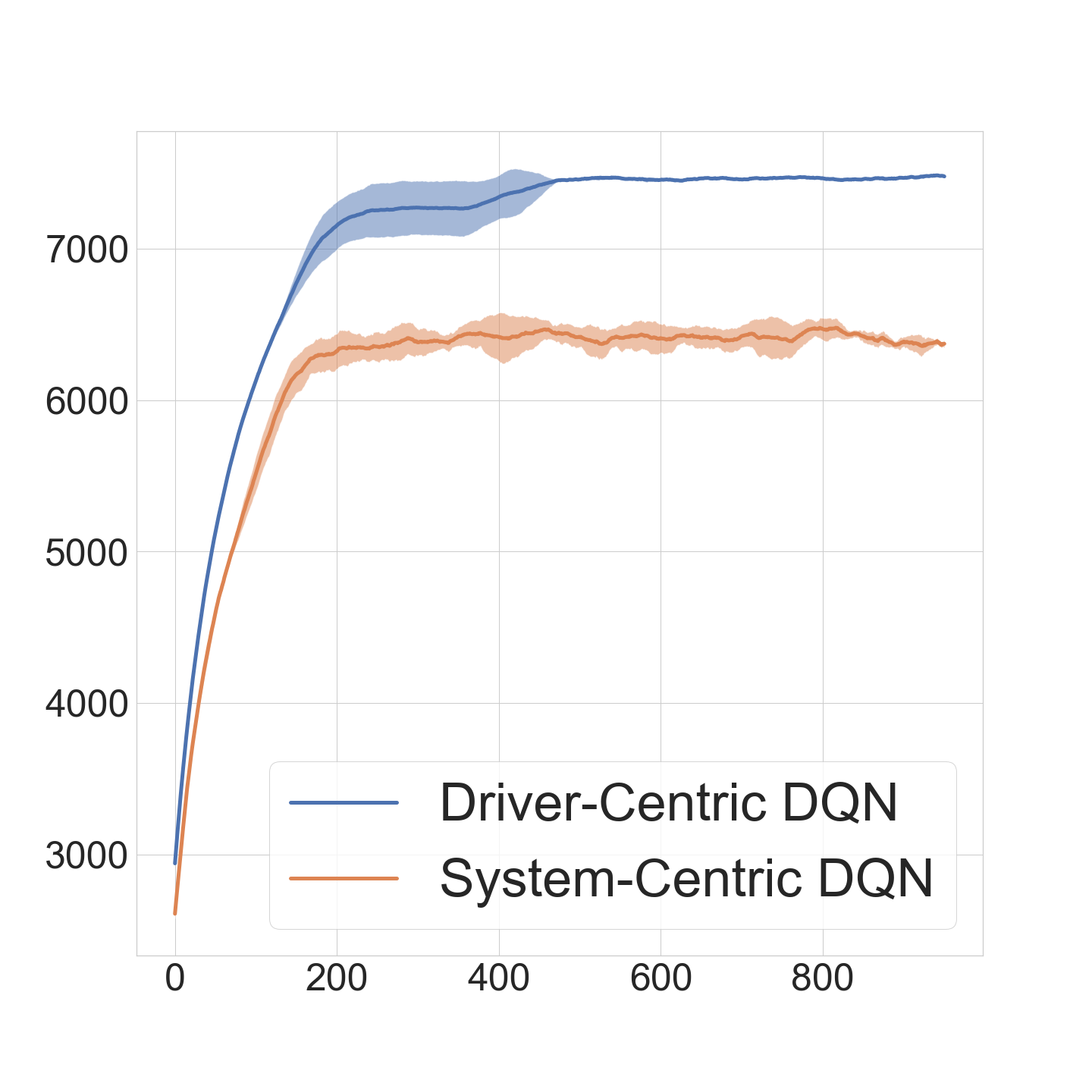} &
        \includegraphics[width=0.22\textwidth,trim={1.8cm 4.0cm 5.0cm 4.0cm},clip]{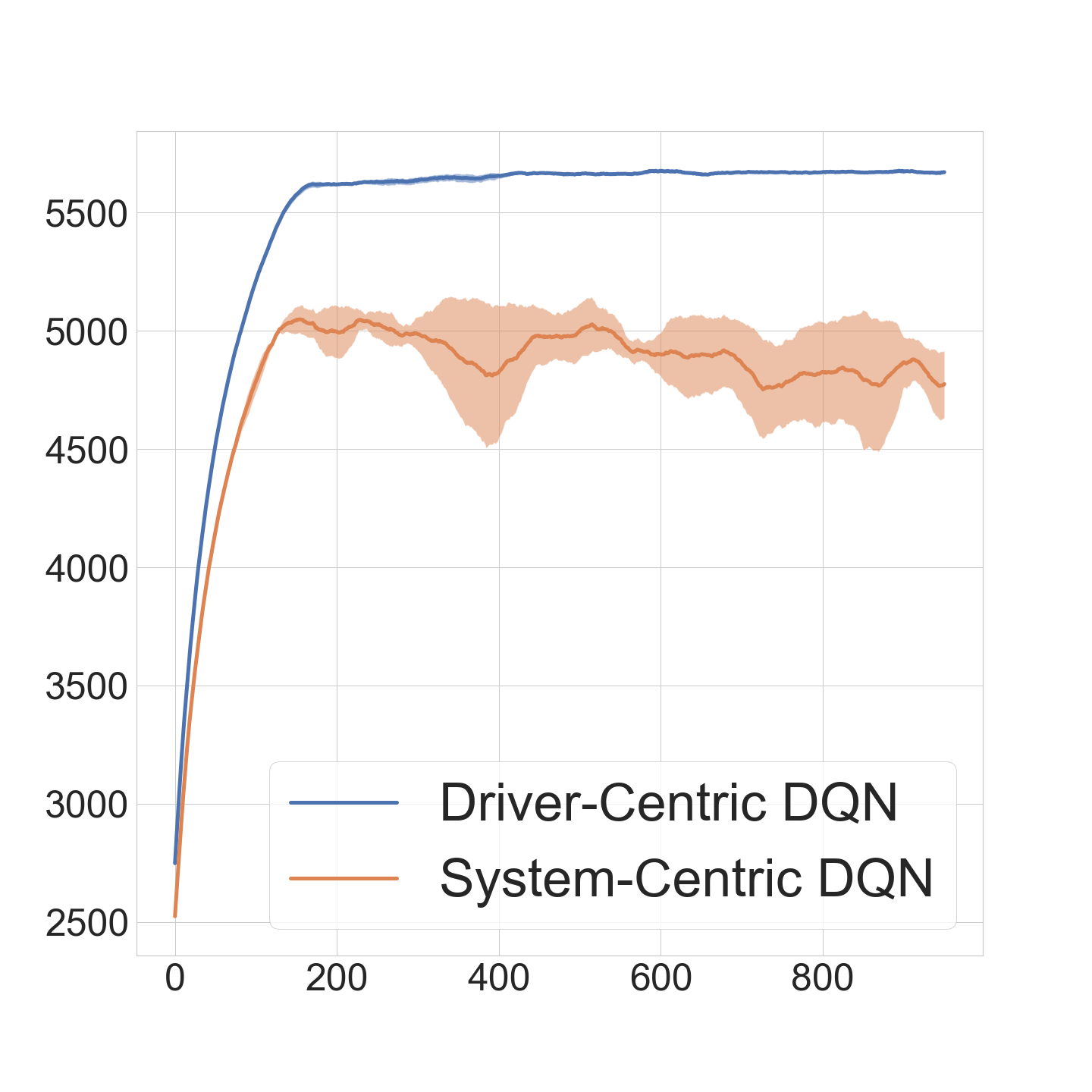} &
        \includegraphics[width=0.22\textwidth,trim={1.8cm 4.0cm 5.0cm 4.0cm},clip]{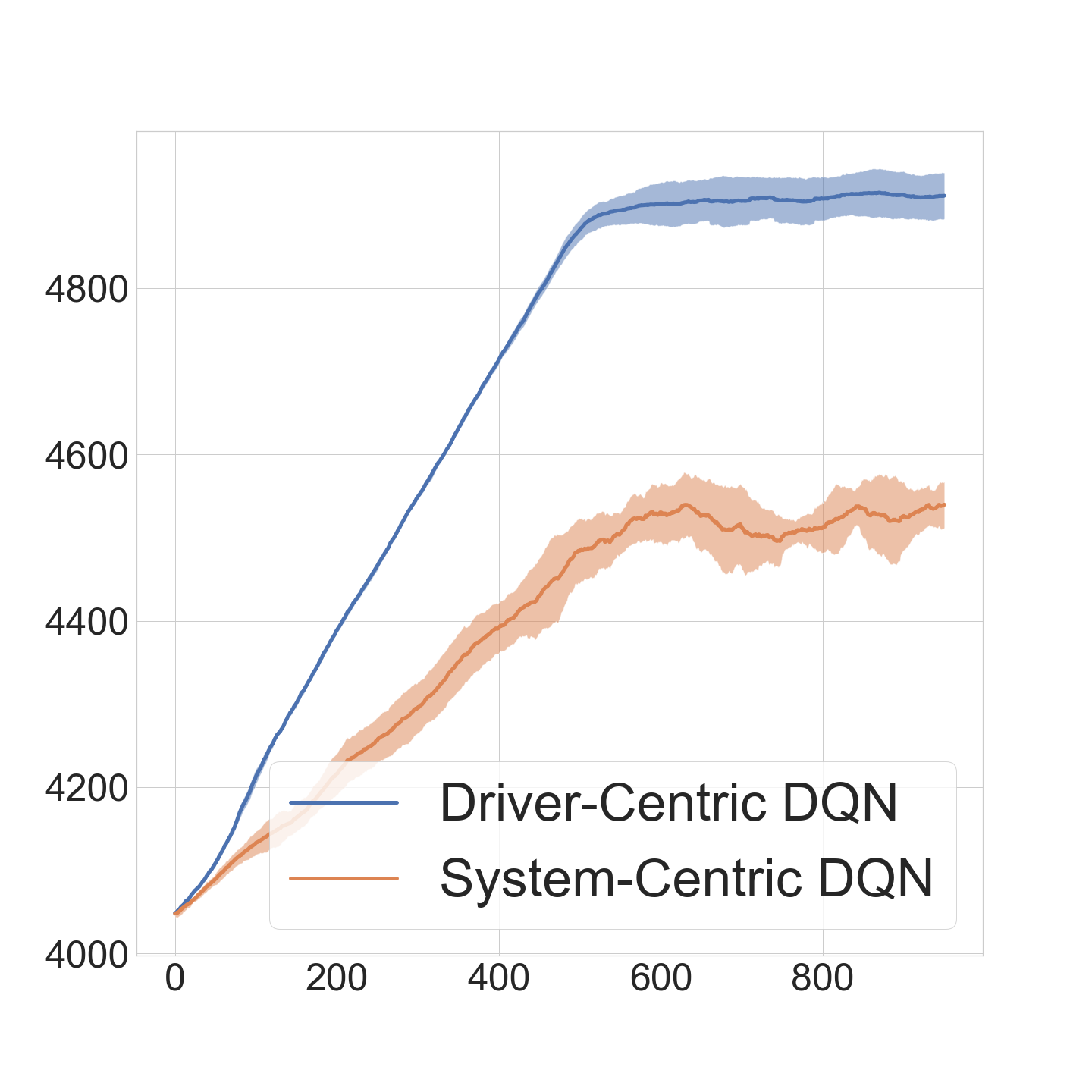} &
        \includegraphics[width=0.22\textwidth,trim={1.8cm 4.0cm 5.0cm 4.0cm},clip]{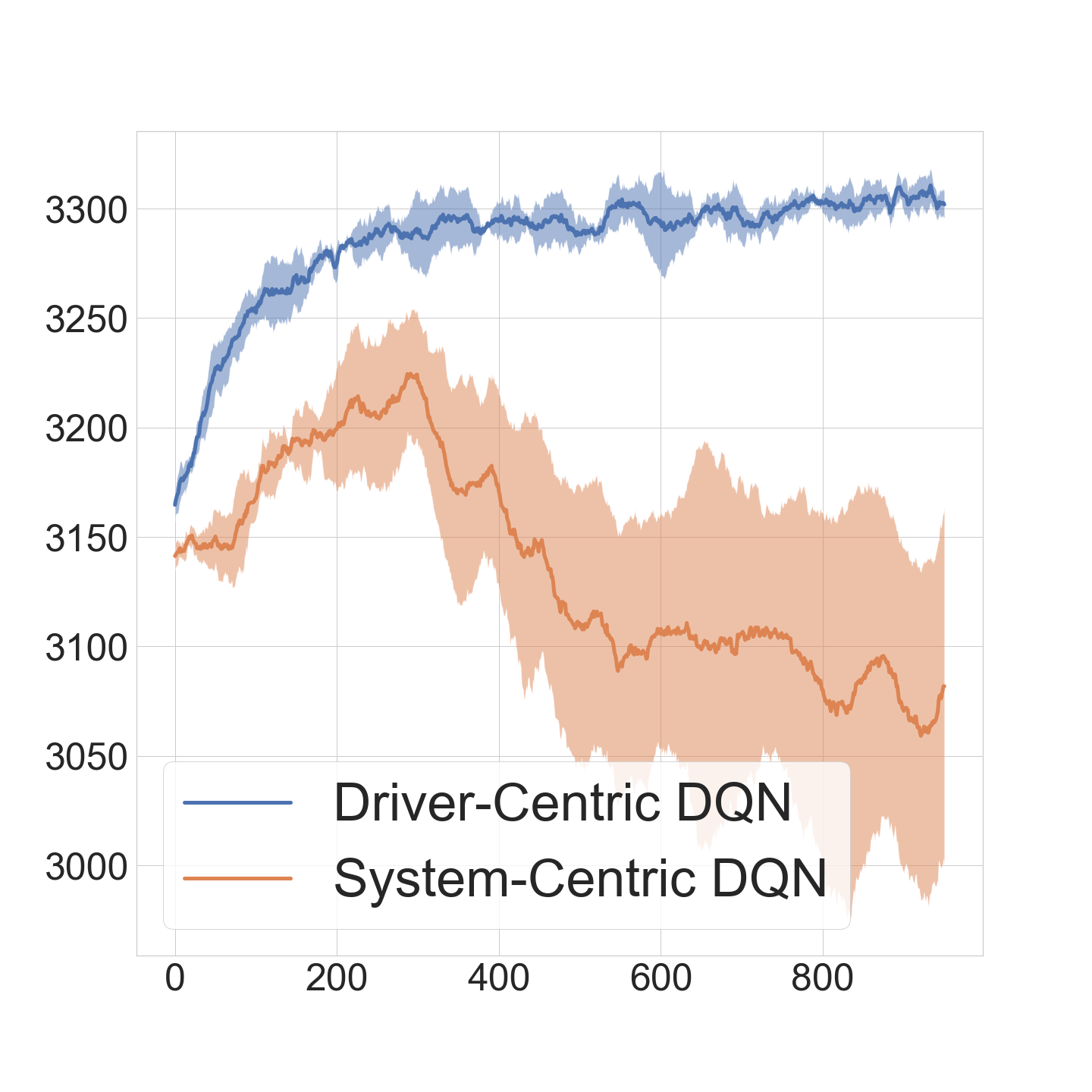}
    \end{tabular}
    \caption{
        \textbf{DQN training curves on the illustrative environments.}
        The curves represent average values across 4 random seeds.
        The shaded area is the standard deviation across the seeds.
        The curves have been smoothed with a sliding window of width 50.
        The vertical axis represents the total driver income.
        The number of training episodes is on the horizontal axis.}
    \label{fig:dqn_training_curves}
\end{figure*}
\begin{figure*}
    \centering
    \begin{tabular}{cccc}
        Hot-Cold High Demand & Hot-Cold Low Demand & Regional High Demand & Regional Low Demand\\
        \includegraphics[width=0.22\textwidth,trim={1.2cm 4.0cm 4.8cm 4.0cm},clip]{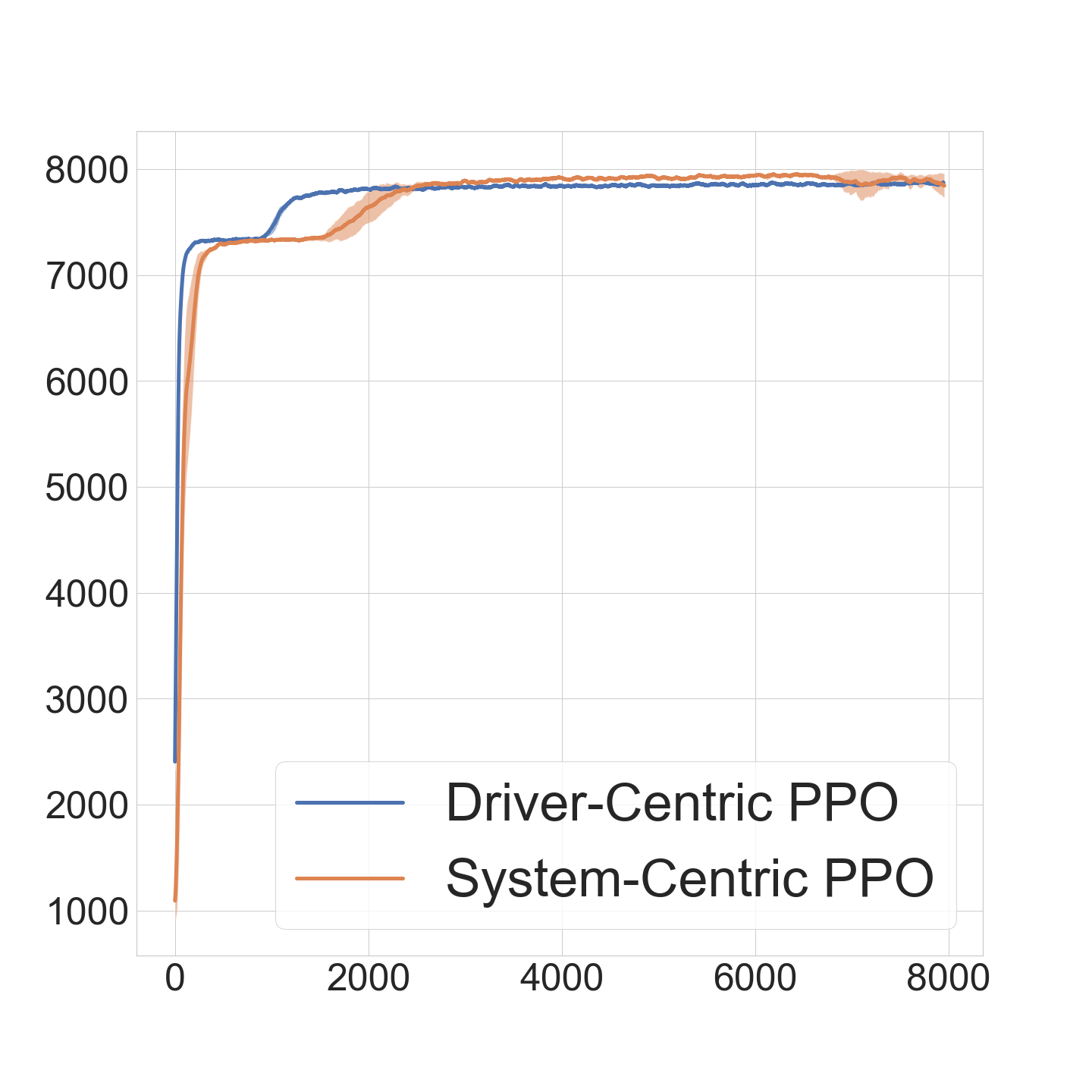} &
        \includegraphics[width=0.22\textwidth,trim={1.2cm 4.0cm 4.8cm 4.0cm},clip]{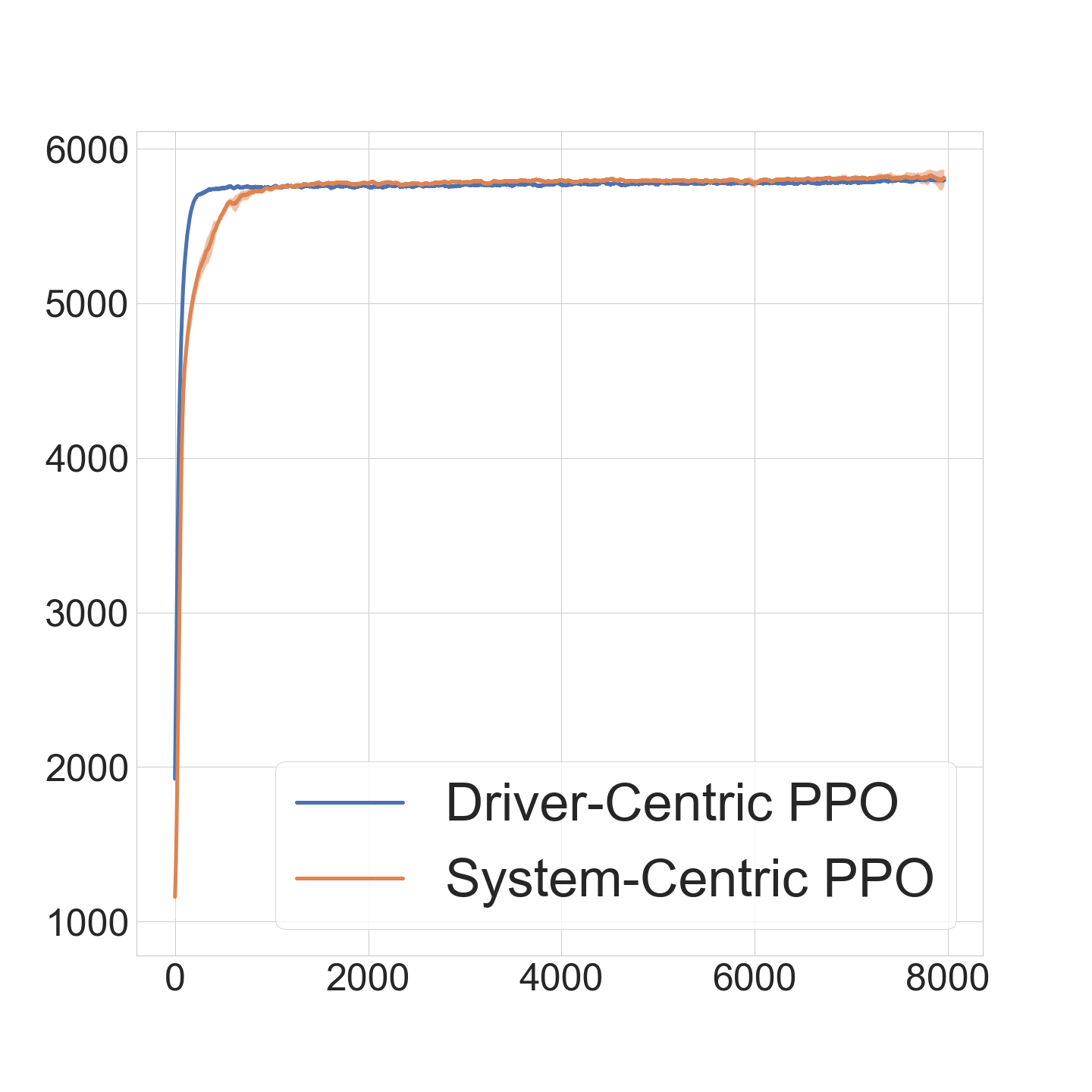} &
        \includegraphics[width=0.22\textwidth,trim={1.2cm 4.0cm 4.8cm 4.0cm},clip]{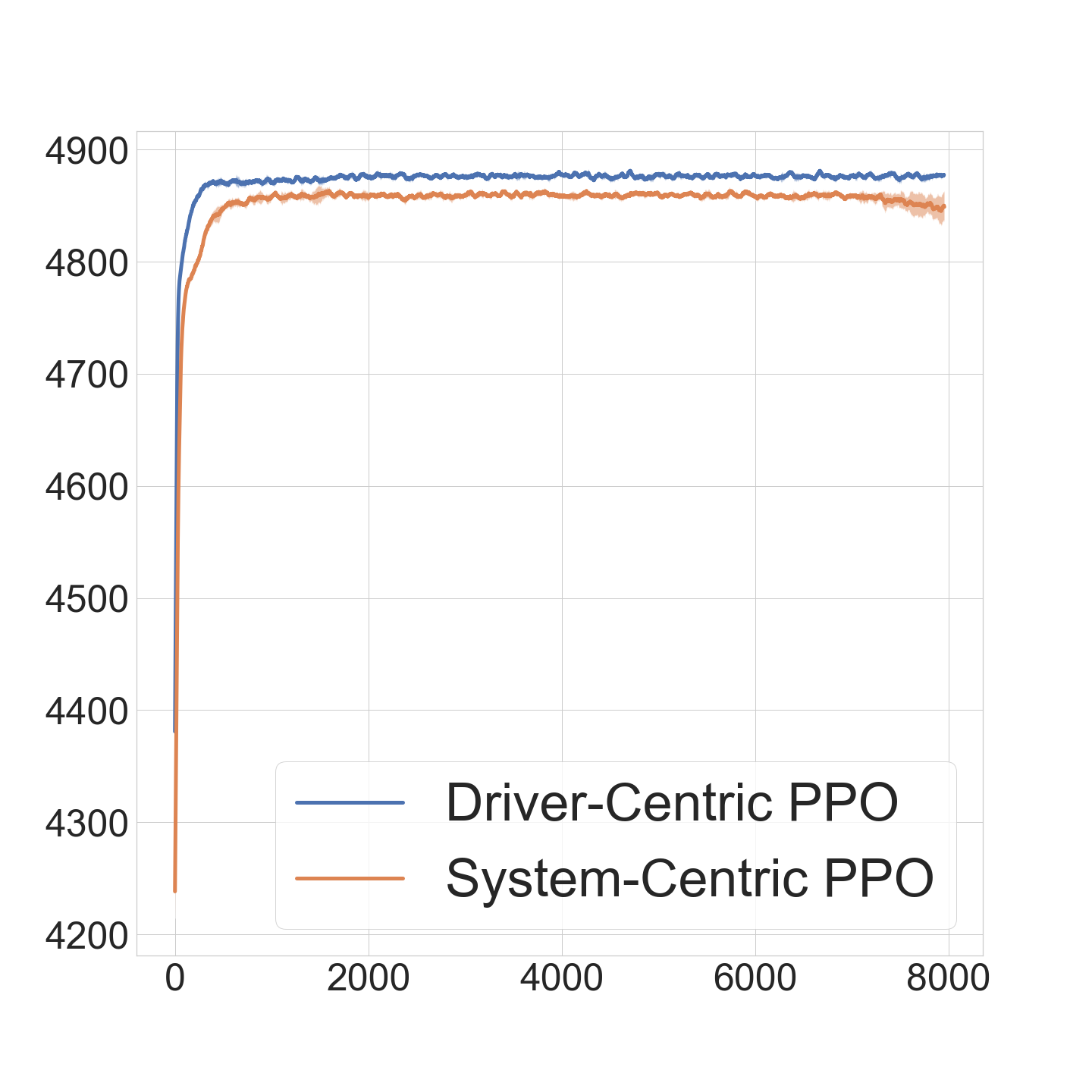} &
        \includegraphics[width=0.22\textwidth,trim={1.2cm 4.0cm 4.8cm 4.0cm},clip]{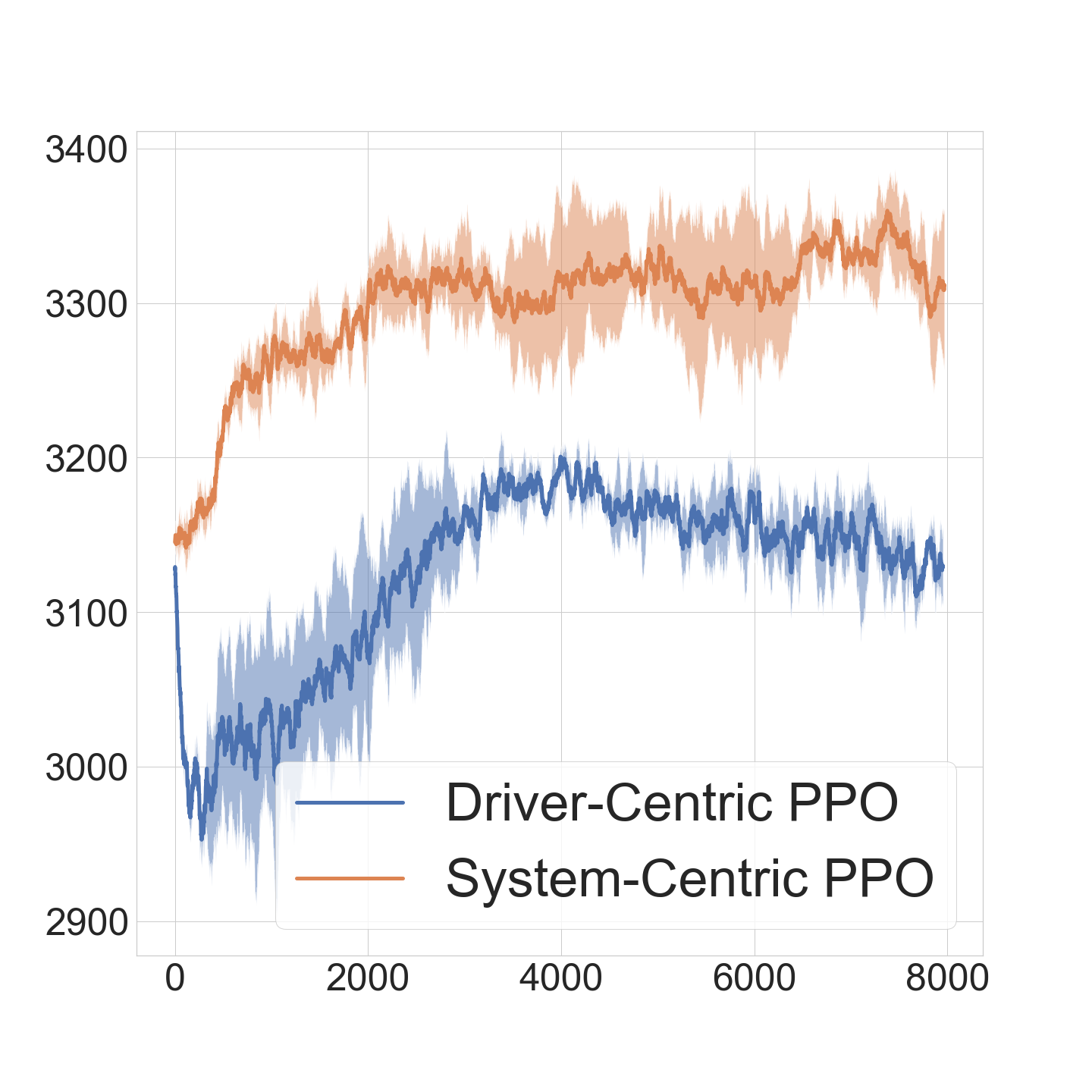}
    \end{tabular}
    \caption{
        \textbf{PPO training curves on the illustrative environments.}
        The curves represent average values across 4 random seeds.
        The shaded area is the standard deviation across the seeds.
        The curves have been smoothed with a sliding window of width 500.
        The vertical axis represents the total driver income.
        The number of update epochs is on the horizontal axis.}
    \label{fig:ppo_training_curves}
\end{figure*}

\begin{figure*}
    \centering
    \begin{tabular}{cccc}
        PPO Historical Statistics & PPO Historical Orders & DQN Historical Orders & \\
        \includegraphics[width=0.22\textwidth,trim={0.3cm 4.0cm 4.9cm 4.0cm},clip]{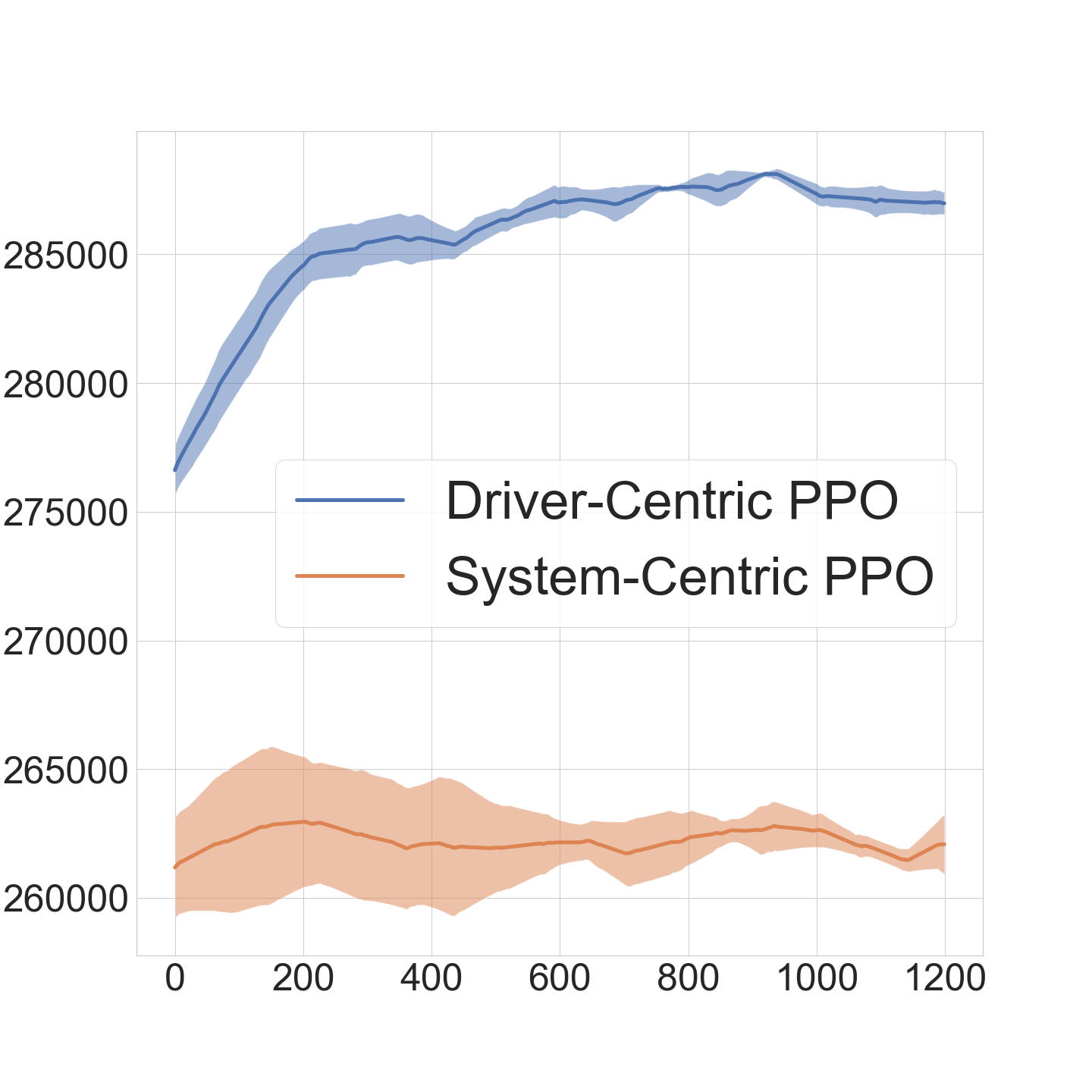} &
        \includegraphics[width=0.22\textwidth,trim={1.2cm 4.0cm 4.5cm 4.0cm},clip]{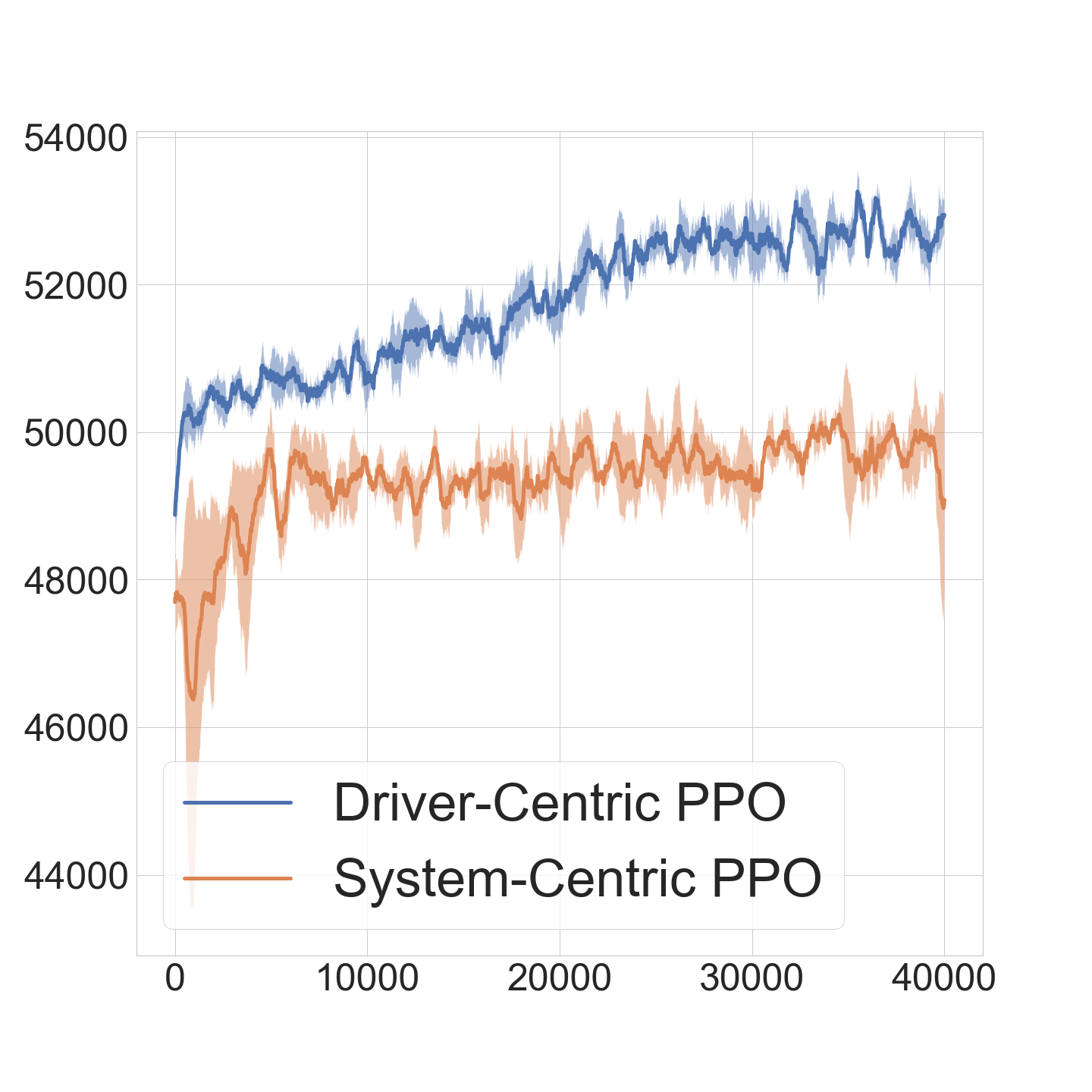} &
        \includegraphics[width=0.22\textwidth,trim={1.2cm 4.0cm 4.8cm 4.0cm},clip]{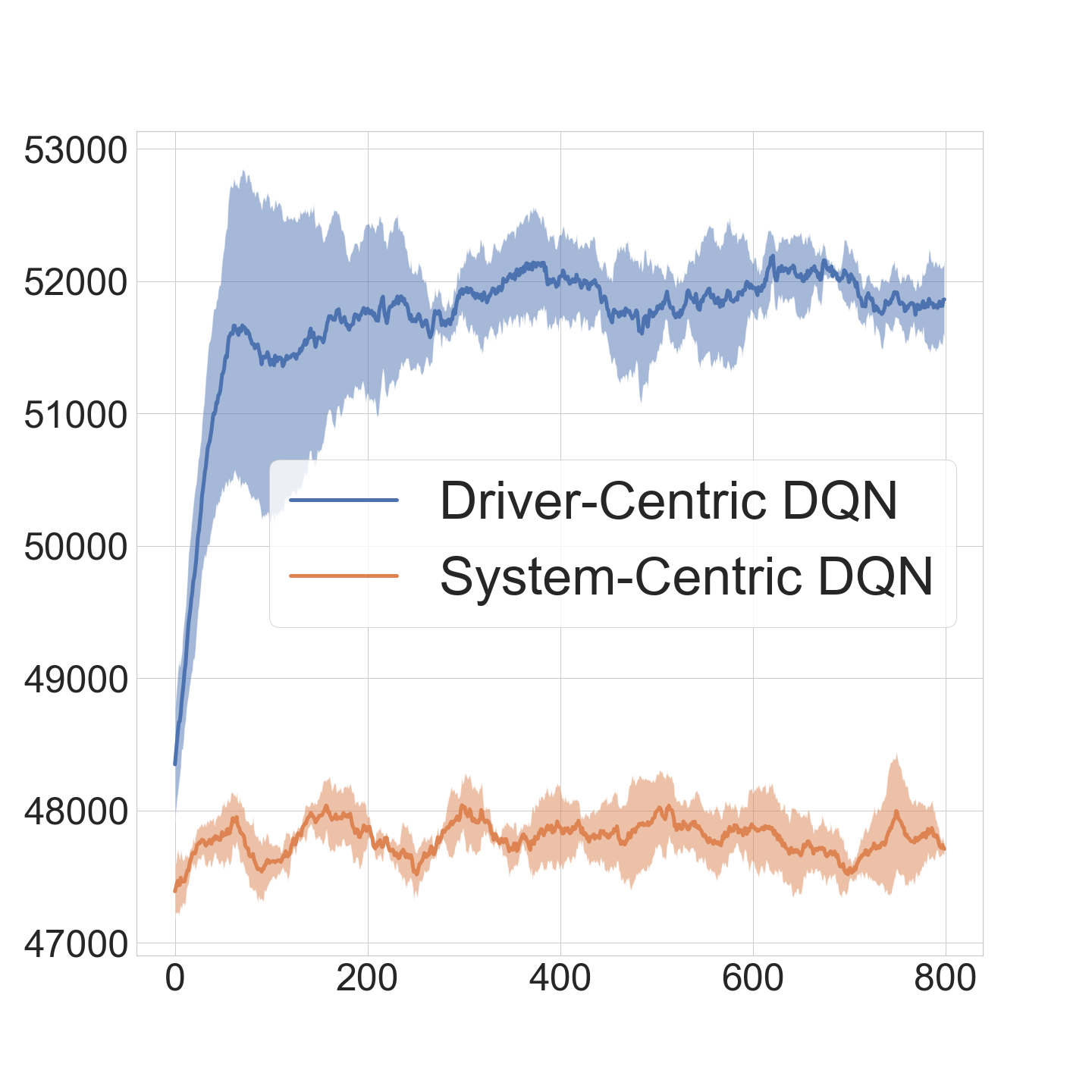} &
    \end{tabular}
    \caption{\textbf{Training curves on real-world data domains.} The curves represent average values across 4 random seeds. The shaded area is the standard deviation across the seeds. The curves have been smoothed with a sliding window of width 500 for PPO and 50 for DQN. The vertical axis represents the total driver income. The horizontal axis represents the number of update epochs for PPO plots and number of training episodes for DQN. Driver-centric rewards learn effectively in these larger scale environments, whereas both algorithms have problems optimizing policies with system-centric rewards.}
    \label{fig:real_data_training_curves}
\end{figure*}